\documentclass{article}
\usepackage{iclr2026_conference,times}

\usepackage{amsmath}
\usepackage{amssymb}
\usepackage{amsfonts}
\usepackage{booktabs}
\usepackage{multirow}
\usepackage{adjustbox}
\usepackage{makecell}
\usepackage{euscript}
\usepackage{xcolor}
\usepackage[theorems,skins]{tcolorbox}
\usepackage{empheq}
\usepackage[leftmargin=1em]{quoting}
\usepackage{placeins}
\usepackage{caption}
\usepackage{subcaption}
\usepackage{listings}
\usepackage{microtype}
\usepackage{algorithm}
\usepackage{algorithmicx}
\usepackage{algpseudocode}
\usepackage[pagebackref,breaklinks,colorlinks]{hyperref}
\usepackage{cleveref}
\usepackage{titletoc}
\usepackage{fontawesome5}
\makeatletter
\renewcommand{\addcontentsline}[3]{%
  \addtocontents{#1}{\protect\contentsline{#2}{#3}{\thepage}{\@currentHref}}}
\makeatother

\usepackage{graphicx}

\DeclareRobustCommand{\coauth}{$^*$}

\DeclareRobustCommand{\iiitdlogo}{
  \raisebox{0.25\height}{\includegraphics[height=1.2ex]{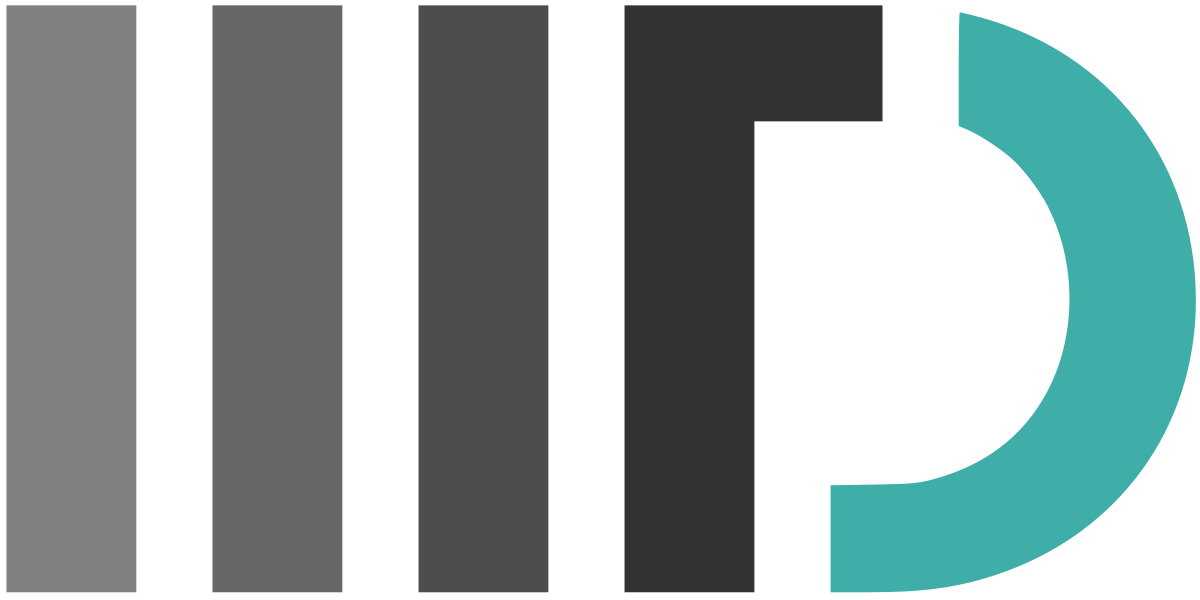}}}

\DeclareRobustCommand{\ublogo}{
  \raisebox{0.25\height}{\includegraphics[height=1.2ex]{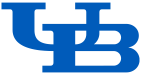}}}

\DeclareRobustCommand{\adobelogo}{
  \raisebox{0.15\height}{\includegraphics[height=2ex]{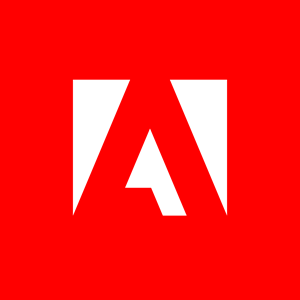}}}

\newcommand\blfootnote[1]{%
  \begingroup
  \renewcommand\thefootnote{}\footnote{#1}%
  \addtocounter{footnote}{-1}%
  \endgroup
}

\newcommand{\auth}[2]{\textbf{#1}~#2}

\usepackage{xspace}
\makeatletter
\DeclareRobustCommand\onedot{\futurelet\@let@token\@onedot}
\def\@onedot{\ifx\@let@token.\else.\null\fi\xspace}

\makeatother

\definecolor{valbest}{HTML}{d9ead3}
\newcommand{\valbest}[1]{\colorbox{valbest}{#1}}
\definecolor{valgood}{HTML}{cfe6ec}
\newcommand{\valgood}[1]{\colorbox{valgood}{#1}}
\definecolor{valbad}{HTML}{f4cccc}

\newcommand{\somesh}[1]{}
\newcommand{\harini}[1]{}
\newcommand{\yaman}[1]{}

\newcommand{\MyModel}{\textsf{ZIPPY}}
\newcommand{\zipp}{{\bf ZIPP}\xspace}
\newcommand{\zipbench}{{\bf ZIP-Bench}\xspace}

\title{Zero-shot Image Personalization From Personas}

\author{%
\parbox{\dimexpr\textwidth-12pt\relax}{\centering\vspace{4mm}
\renewcommand{\arraystretch}{1.35}
\setlength{\tabcolsep}{4pt}
\begin{tabular}{ccc}
  \auth{Harini S I\coauth}{\adobelogo} &
  \auth{Somesh Singh\coauth}{\adobelogo~\ublogo~\iiitdlogo} &
  \auth{Yaman Kumar Singla}{\adobelogo} \\[1mm]
  \multicolumn{3}{c}{%
    \auth{David Doermann}{\ublogo}\qquad
    \auth{Rajiv Ratn Shah}{\iiitdlogo}%
  } \\[3.5mm]
\end{tabular}\\[1mm]
{\adobelogo~Adobe Media and Data Science Research (MDSR)\\[1mm]
 \iiitdlogo~IIIT-Delhi,\quad\ublogo~SUNY at Buffalo \\}
 \faEnvelope\ \texttt{\href{mailto:behavior-in-the-wild@googlegroups.com}{behavior-in-the-wild@googlegroups.com}}
}%
}

\iclrfinalcopy
\begin{document}

\maketitle
\pagestyle{fancy}
\fancyhf{}
\fancyhead[L]{%
  \raisebox{0.01\height}{\includegraphics[height=2ex]{images/adobe-logo.png}}%
  \hspace{0.4em}Adobe Media and Data Science Research%
}
\renewcommand{\headrulewidth}{0.4pt}
\renewcommand{\footrulewidth}{0pt}

\begin{NoHyper}
  \blfootnote{\coauth~Equal contribution.}
\end{NoHyper}

\vspace*{-0.73cm}
\noindent\begin{minipage}{\textwidth}
\centering
\includegraphics[width=\textwidth]{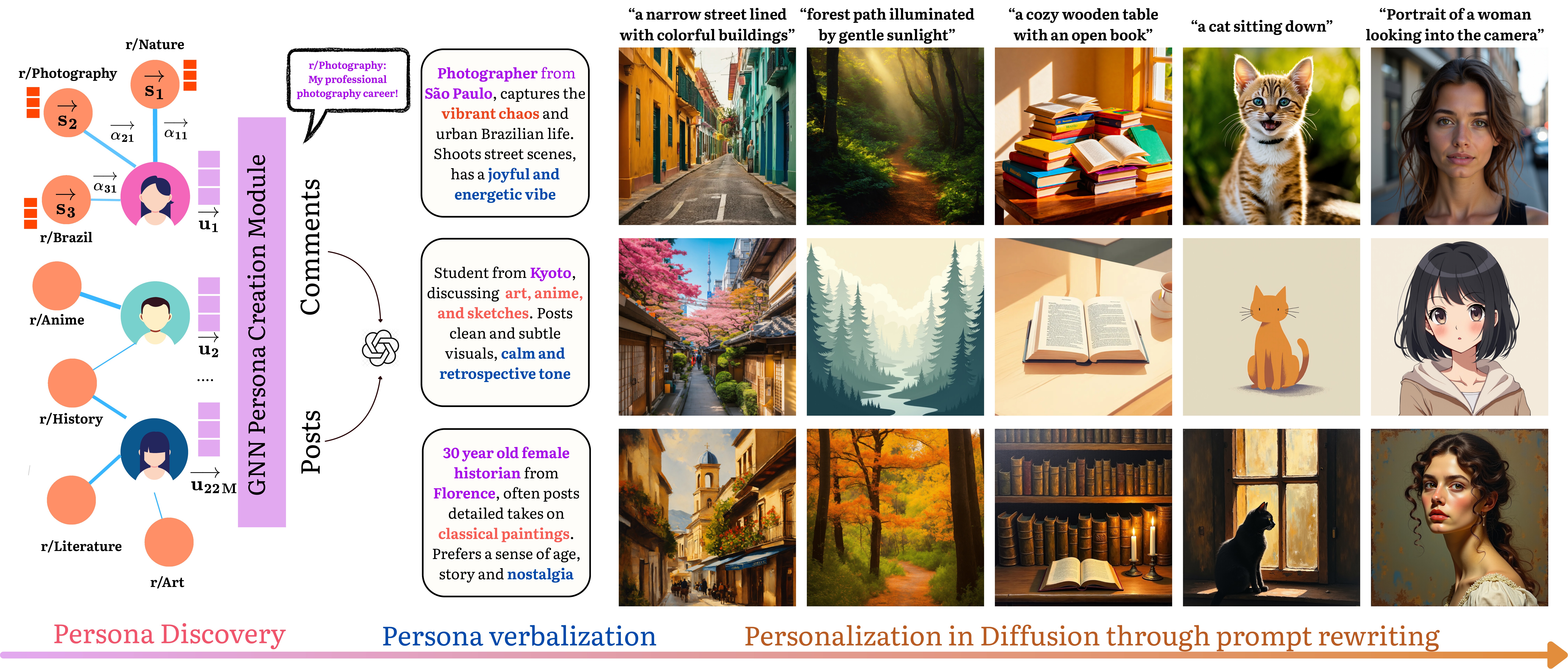}
\captionsetup{hypcap=false}%
\captionof{figure}{\textbf{Overview of our zero-shot personalization framework.}
A graph attention network trained on the user--subreddit bipartite graph ranks subreddits
by attention score to construct natural language \textbf{persona verbalizations} encoding attributes like: 
\textcolor{violet}{\textbf{demographic attributes}}, \textcolor{red}{\textbf{interests}},
and \textcolor{blue}{\textbf{affective traits}}.
These personas condition an LLM to rewrite base prompts for \textbf{zero-shot personalization},
producing images aligned with individual user preferences.}
\label{fig:teaser}
\end{minipage}

\begin{abstract}

Text-to-image diffusion models are increasingly deployed in creative contexts, yet remain impersonal—optimized for aggregate aesthetics rather than individual taste. Human preferences are inherently pluralistic: one user who favors muted, nostalgic portraits may prefer vibrant compositions for street photography, while another gravitates toward dreamy, overexposed film aesthetics. Existing personalization methods require dense interaction histories or per-user fine-tuning, failing in cold-start settings and collapsing each user's context-dependent preferences into a single static representation. We introduce \emph{zero-shot image personalization from personas} (\zipp), a paradigm that conditions image generation on natural-language personas—concise descriptors of a user's identity, interests, and aesthetic sensibilities—without any user-specific data or weight updates. \zipp uses an LLM in a roleplay setting to rewrite input prompts from the perspective of a given persona, steering diffusion models toward personalized outputs. To mine personas at scale, we develop an inductive Graph Attention Network over a 22M-user Reddit interaction graph with dual contrastive objectives that align graph structure with users' visual behavior, and verbalize learned representations into coherent natural-language personas via an MLLM. We further introduce \zipbench, the first zero-shot image personalization benchmark, pairing 1.5K users with graph-mined personas and 40K generated images. Across four benchmarks and 14 LLMs spanning five model families, persona conditioning yields consistent improvements in both zero-shot and few-shot settings, with frontier models achieving the strongest gains (13--20\%). In the few-shot setting, \zipp matches or exceeds fine-tuned baselines requiring per-user adapters trained on 100+ examples. Unlike baselines that collapse preferences into a fixed style, \zipp preserves intra-user preference diversity, achieving the lowest distributional divergence from users' true preference distributions (CMMD 0.16 vs.\ 0.55 for fine-tuned alternatives). IPF-normalized evaluation against global population demographics further reveals that existing methods exhibit substantial bias toward narrow subpopulations, which persona conditioning significantly mitigates. A human evaluation confirms these findings: \zipp achieves a 79\% win rate over generic generation and outperforms all fine-tuned baselines (58--65\% win rate) without any user-specific training.

\end{abstract}

\section{Introduction}

\label{sec:intro}
\begin{quoting}
``\textit{We must design for the way people behave}" -- Don Norman
\end{quoting}
\vspace{-0.4em}

Personalization has emerged as a foundational challenge in generative modeling. State-of-the-art text-to-image diffusion models such as DALL·E \cite{dalle2021}, NanoBanana \cite{nanobanana2025}, Firefly \cite{firefly2024}, and Qwen-Image \cite{qwenimage2025} produce images of remarkable fidelity and semantic alignment, yet their outputs remain impersonal: optimized for aggregate aesthetic preferences that reflect an implicit ``average user.'' In practice, the same prompt elicits vastly different expectations from different people: a photographer may desire film grain and natural light, while a digital artist may prefer vibrant palettes and geometric compositions. Even a single user expresses different visual preferences depending on context: muted tones for documentary and saturated color for social media. As generative systems increasingly mediate creativity and communication, aligning their outputs with both the variation \emph{across} users and the variation \emph{within} a single user's contexts constitutes a central open problem.

In principle a generative model trained across the global population and all possible contexts could solve this. In practice, this is infeasible, collecting dense, user preference histories is expensive and inherently non-scalable, with the largest publicly available datasets accounting for $<$0.001\% of the global population. Moreover preferences evolve over time, making static preference profiles brittle. Further, existing preference annotation and collection pipelines disproportionately sample from a narrow slice of the global population, leaving a majority of potential users unrepresented. As a result, both the models and the benchmarks that evaluate them encode a narrow band of aggregate aesthetic norms that need to be continuously maintained. 

Current methods attempt to solve this challenge by modeling user preferences through shallow user feedback including: pair wise preferences \cite{von2024fabric,salehi2024vipervisualpersonalizationgenerative}, prompt histories \cite{chen2024tailored,kim2025draw}, or detailed comments \cite{salehi2024vipervisualpersonalizationgenerative}. First, these methods fail to personalize without interaction data (the cold-start problem) or transfer across users. Further, user preferences shift across contexts (see \cref{fig:pluralism}), as a result these methods collapse each user into a single, static representation, discarding the pluralistic and context-dependent nature of human preference (see \S\ref{ssec:pluralism}). Consequently, personalization systems must not only learn from individual user data, but also generalize effectively to users with little or no explicit feedback.

To solve this challenge, we propose personalizing image generation by conditioning generative models on natural language personas. Personas capture rich demographic, social, cultural, and behavioral contexts which influence their visual preferences. With this we enable a new paradigm called \zipp (Zero-shot Image Personalization from Personas) solving the cold start problem. Concretely, ZIPP conditions image generation by rewriting input prompts, through an LLM roleplaying the given persona. For example, a persona describing a Florence-based historian in \cref{fig:teaser} emphasizes classical art, historical atmosphere, and narrative depth; when applied to a generic prompt such as ``a cozy wooden table with an open book" the LLM appends `weathered materials, muted tones, and historical' to it. This rewritten prompt then conditions a downstream diffusion model, steering generation toward images aligned with the persona's preferences.

\begin{figure*}[t]
\centering
\includegraphics[width=1\textwidth]{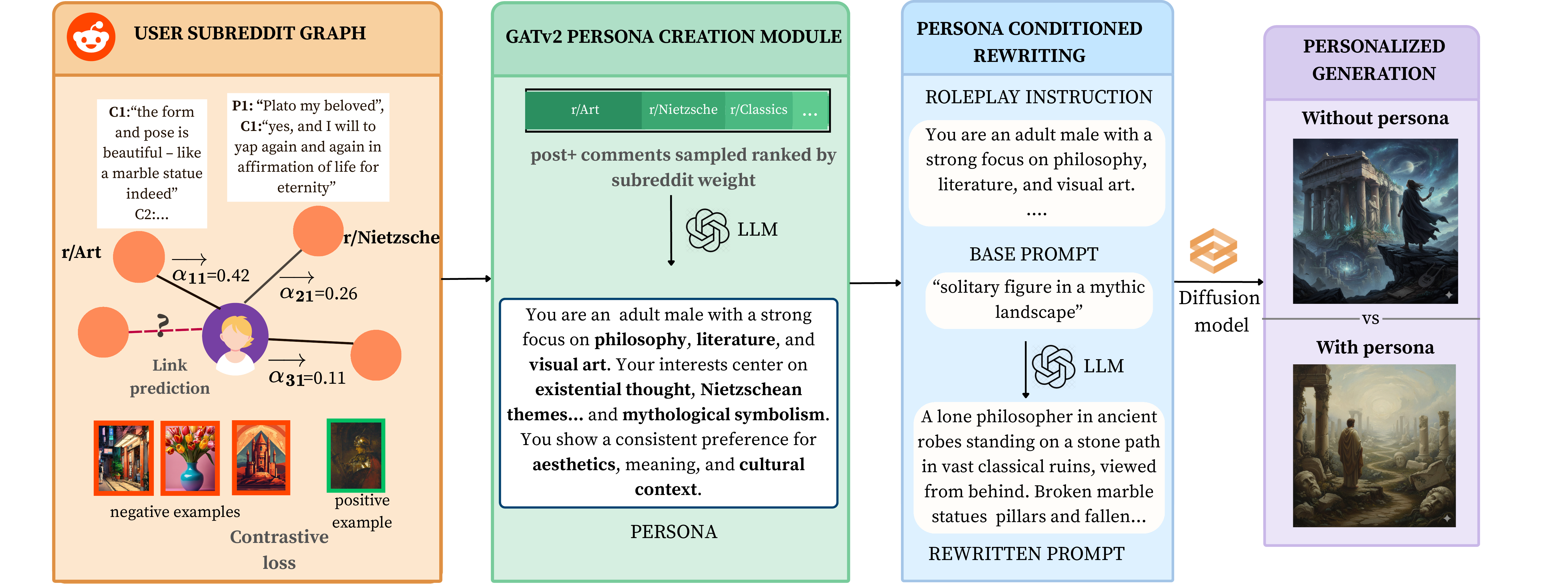}
\vspace{-1.7em}
\caption{Overview of methodology.}
\label{fig_methodology}
\end{figure*}

This paradigm offers several advantages. First, it generalizes to users for whom no prior feedback is available, addressing the cold-start problem. Second, persona-conditioned prompts are explicit and editable, providing transparency, auditability and user control. Third, the approach transfers across image generation models without retraining. Finally, LLMs can simulate diverse behaviors and preferences through personas ~\cite{du2025twinvoice} and offer a higher coverage of the global population. Concretely, our contributions are fourfold:

\label{ref:our_contributios}
\begin{itemize}
    \item \textbf{Paradigm:} We introduce \emph{zero-shot image personalization}, a paradigm that enables image personalization without explicit image preference data. \MyModel, an LLM-based framework rewrites input prompts conditioned on user persona. We achieve strong and consistent additive gains, i.e. they show improvement in both zero and few-shot settings compared to state-of-the art methods (see \Cref{tab:main_results}).
    \item \textbf{Method:} We develop a novel Graph Attention-based inductive framework to mine personas from large-scale social graphs that are predictive of their visual preferences illustrated in \Cref{fig_methodology}.
    \item \textbf{Data:} We also introduce \textbf{\zipbench}, the first benchmark for zero-shot image personalization, comprising 1.5K users, their detailed and predictive natural-language personas, and 40K generated images (see \Cref{sec:data}), constructed using our graph neural network. We benchmark and analyze multiple open and frontier models and show consistent improvement of 3-25\%. 
    \item \textbf{Pluralism and Demographic Alignment:} We identify existing inter- and intra-user personalization challenges in current benchmarks and methods. Existing benchmarks are largely biased towards subpopulations and methods fail to personalize across diverse contexts. We propose methods to evaluate and mitigate these challenges in \Cref{ssec:pluralism}
\end{itemize}

\section{Background \& Related Work}

A persona typically refers to a textual representation of an individual's consistent traits, preferences, or style. Our work treats personas as zero-shot conditioning vectors for personalization. We review two areas: (1) personalization methods for text-to-image models, (2) user behavior modeling and persona creation frameworks. We position our work at the intersection of their work.

\vspace{-0.8em}
\paragraph{Image Personalization Methods.}
Personalized image generation adapts text-to-image models to reflect individual visual preferences, spanning aesthetic styles, compositional patterns, subject matter affinities, and semantic interpretations that vary across users. Existing approaches fall into two categories: \textit{subject-driven personalization}, which reproduces specific concepts or styles~\cite{ruiz2023dreambooth, gal2022image, sohn2023styledrop, park2025steering}, and \textit{user-based personalization}, which models broader user taste from behavioral or interaction data. User-based approaches~\cite{von2024fabric, salehi2024vipervisualpersonalizationgenerative, chen2024tailored, xu2025personalized, kim2025draw} typically learn preferences from explicit feedback (liked/disliked images), historical prompts, or prompt–image pairs.  
However, these methods face key limitations: (1) they require dense per-user histories (10–100+ samples) or degrade with sparse or abstract inputs, (2) depend on non-scalable, and costly fine-tuning or retrieval pipelines, (3) fail to generalize across contexts for a user, and (4) fail to personalize to unseen users.
For example, ViPer~\cite{salehi2024vipervisualpersonalizationgenerative} requires 8–20 detailed annotated images and produces context-insensitive prompts (e.g., applying ``vibrant colors" universally), while Tailored Vision~\cite{chen2024tailored} deteriorates when conditioned on more than three previous prompts.  

Our approach provides a zero-shot, language-only alternative where natural-language personas derived without explicit annotations act as interpretable conditioning signals for personalized image generation, eliminating the need for per-user histories, feedback, or fine-tuning.

\vspace{-0.5em}
\paragraph{Persona Creation and Behavioral Modeling.}
Modeling user preferences through personas has long been central to personalization and HCI research. Recent work scales this process using large language models (LLMs). Persona Hub~\cite{ge2024scaling} and PersonaCraft~\cite{jung2025personacraft} generate diverse persona corpora from web and survey data, while others~\cite{shin2024understanding} employ human–LLM workflows for fine-grained clustering. Role-play prompting further shows that LLMs can internalize persona traits across dialogue~\cite{ng2024well, tang2023enhancing}, recommendation~\cite{yang2023palrpersonalizationawarellms}, search~\cite{zhou2024cognitivepersonalizedsearchintegrating}, and reasoning~\cite{kong2024betterzeroshotreasoningroleplay} tasks.

In multimodal modeling, Behavior-LLaVA~\cite{singh2025teaching} shows that user comments on visual content capture perceptual and emotional salience. While studied in the visual domain, this insight generalizes more broadly, as user commentary often encodes stable personal attributes such as demographics, topical interests, communication style, and aesthetic preferences. However, existing persona datasets remain either synthetic (Persona Hub) or survey-based (PersonaCraft), \somesh{This seems a little off}and behavioral traces are typically used to evaluate learned personas rather than to construct them.
\somesh{What Gap?}
Our approach bridges these fields by unifying persona mining and image personalization within a single inductive framework. We extract personas from large-scale social graphs by modeling users' network, activity, and behavior (posts and comments) and use these personas as conditioning signals for personalized image generation. Unlike prior methods that rely on dense preference histories or per-user fine-tuning, we demonstrate that natural-language personas alone can effectively guide image generation across entirely new users, contexts, and diffusion model families.

\section{Methodology}

\label{sec:data}
\vspace{-0.5em}\subsection{Dataset Curation for Personas}
Existing image personalization benchmarks and datasets suffer from two critical limitations: (1) they capture preferences from narrow, self-selected user demographics (e.g., creative hobbyists on Pick-a-Pic, expert annotators on HPD-v2), limiting generalization to diverse user populations; and (2) they lack explicit user identity representations (personas). These inhibit us from evaluating and optimizing accurately.

To address this, we construct two complementary datasets that enable large-scale persona mining and zero-shot personalization evaluation across diverse user populations.

\vspace{-1em}\paragraph{Reddit Interactions}
Persona mining requires large-scale behavioral data that reveals authentic visual preferences across heterogeneous users. Unlike survey-based persona datasets (PersonaCraft\cite{jung2025personacraft}) or synthetically generated personas  (Persona Hub \cite{ge2024scaling}), we ground personas in real interaction traces from Reddit\footnote{\url{https://reddit.com/}}, a platform with 400M+ monthly active users spanning diverse demographics and visual preferences.
Reddit's community structure provides a natural substrate for this task: users engage with topic-specific subreddits (e.g., r/AiArt) by commenting or posting, revealing their latent preferences, and in the process creating a bipartite graph of subreddits and users. We aggregate seven years of activity (2015–2022) from 23M users across 40K subreddits, capturing 682M edges of unique subreddit and user interactions. This graph structure provides strong inductive biases for persona discovery through interest neighbourhoods. Beyond Reddit, this graph-based approach can be generalized to any platform with a user-community structure. We focus on Reddit for its scale, API accessibility, and diverse visual communities spanning photography, generative art, and niche interests. In \cref{sec:persona_mining} we describe  how we discover natural language personas (e.g. ``student from Kyoto") from this graph.

\vspace{-1em}\paragraph{ZIP Bench}
While the Reddit graph enables persona mining at scale, benchmarking zero-shot image personalization requires users with both rich interaction histories and visual preferences or historical generations. We focus on Civitai\footnote{\url{https://civitai.com/}}, one of the most popular open-source AI-generated content (AIGC) platforms where users create, customize, and share diffusion models and generated images. Crucially, Civitai's public API provides detailed metadata (prompts, model versions, NSFW classifications), and many platform users maintain linked Reddit accounts, enabling us to pair graph-mined personas with real-world generation histories.

Utilizing this cross-profile linking, we identify 1.5K Civitai users active on AI art subreddits (r/civitai, r/AIArt), extracting their complete Reddit history (15K posts, 183K comments from 2015-25) to mine personas from the global graph, alongside 40K generated images with prompts from their Civitai profiles. Unlike previous personalization benchmarks~\cite{chen2024tailored, kirstain2023pick}, ZIP Bench uniquely combines explicit natural-language personas with generation histories, enabling the first systematic evaluation of whether and how well persona descriptions alone can reliably guide personalization without per-user fine-tuning (\cref{sec:exp_results}).

\label{sec:persona_mining}
We now detail our two-stage pipeline to (1) learn distinct user representations from large social networks and (2) use them to create personas which are indicative of their implicit visual behavior.
\textit{persona} is a concise textual descriptor of a user's interests, aesthetic 
tendencies, and engagement patterns (e.g., ``anime enthusiast favoring pastel palettes 
and character-focused compositions")—serving as an interpretable, zero-shot conditioning 
signal for image personalization. Our goal is to mine such personas at scale from the 
Reddit interaction graph, ensuring they (1) reflect authentic behavioral patterns rather 
than self-reported surveys, (2) cluster users by visual preference similarity, and 
(3) generalize to unseen users without requiring per-user annotation.

\paragraph{Graph Setup:} We model the Reddit interactions as a bipartite graph
$\mathcal{G}=(\mathcal{U},\mathcal{S},\mathcal{E})$ where users $\mathcal{U}$ and 
subreddits $\mathcal{S}$ form distinct node types, connected by directed edges 
$\mathcal{E}\subseteq \mathcal{U}\times \mathcal{S}$ representing posting and 
commenting activity. For each interaction $(u,s)\in\mathcal{E}$, we aggregate activity 
counts $t_{us}$ (posts + comments) and compute node degrees $c_u=\sum_{s} t_{us}$, 
$c_s=\sum_{u} t_{us}$. Akin to social networks, we observe that Reddit-Interactions has (1) long-tailed activity distributions (12k subreddits contribute to only 0.1\% of the total activity) and (2) popularity bias (popular subreddits and power-users have a very high activity). To alleviate this, we remove the bottom 12k subreddits and assign the edge weight to be log and degree normalized activity
$w_{us}=\log(1+t_{us})/\sqrt{c_u c_s+\varepsilon}$ following PinSAGE's \cite{Ying_2018}
PMI-style normalization (\cref{fig:deg-subreddits,fig:deg-users} shows a normal distribution on the log/log scale supporting this). The resulting graph consists of 32k subreddits, 681M Edges, and 22.8M edges. Additional 
statistics, edge weighting schemes, and ablations are in 
\cref{sec:supp-gnn}. To cluster users or communities at this scale using traditional graph algorithms is infeasible; therefore, we use graph neural networks to learn these representations.
\vspace{-0.5em}\paragraph{Learning User Representations}
It is essential to learn user representations inductively from the graph structure, enabling generalization to new users at inference without requiring retraining. Since there are only 32k subreddits and we have their detailed descriptions, we initialize the nodes with pretrained text embeddings $\mathbf{x}_s \in 
\mathbb{R}^{3072}$ (OpenAI text-embedding-3-large ~\cite{openai_text_embedding_3_large_2024}) of these descriptions and user nodes with a small gaussian noise, the nodes and edges are projected into hidden states $\mathbf{h}_u,\mathbf{h}_s,\mathbf{h}_{e}$ respectively. We employ two GATv2 layers \cite{brody2022attentivegraphattentionnetworks} with 4 attention heads each, aggregating information from subreddits to users via attention mechanism, the resulting user embedding $\mathbf{h_u}$ is the standard multi head attention, where the output of each attention head $\mathbf{h}_u^{'(l)}$ is defined as

\begin{equation}
    \quad \mathbf{h'}^{(\ell)}_u = \sum_{s \in \mathcal{N}(u)} \alpha^{(\ell)}_{us} 
    \mathbf{W}_v^{(\ell)} \mathbf{h}_*^{(\ell)}
    \label{eq:GATConv2}
\end{equation}

where $\alpha^{(\ell)}_{us}$ is the graph attention for the head $\ell$, between the user $u$ and its neighbors $s \in \mathcal{N}(u)$ and $\mathbf{W},\mathbf{h}_*$ are the parameters and concatenated hidden states $\mathbf{h}_u,\mathbf{h}_s,\mathbf{h}_{e}$. We use the standard LeakyReLU (slope 0.2) activations, apply LayerNorm before each attention block, and L2-normalize the node embeddings $\mathbf{z}_u = \mathbf{h'}_u / \|\mathbf{h'}_u\|_2$ (and similarly $\mathbf{z}_s$ for subreddits $\mathbf{h}_s$) to maintain cosine geometry for our objectives that we define below

\textbf{Dual contrastive objectives.} We supervise the embedding space through the self-supervised link prediction task, we sample edges uniformly across subreddits and users, and use the InfoNCE loss to contrastively align the user embeddings $\mathbf{z}_u$ with their neighbors' embeddings $\mathbf{z}_s$ (negatives are sampled from close neighbors).
Let $\mathcal{B} = \{(u_i, s_i)\}_{i=1}^B$ be a batch of edges, the objective is to minimize the loss 
\vspace{-0.8em}
\[
\mathcal{L}_{u,s} = \frac{1}{B} \sum_{i=1}^{B} 
    \mathrm{CE}( 
         \mathbf{z}_{u_i}^\top [\mathbf{z}_{s_1}, \ldots, \mathbf{z}_{s_B}], 
        i 
    \bigr),
\]
\label{eq:loss-us}
\vspace{-0.8em}

where $\mathrm{CE}(\cdot, i)$ is cross-entropy treating $i$ as the positive index 
and all other subreddits in the batch as negatives similar to CLIP \cite{radford2021learning}. 

To explicitly model visual behavior signals we add a 
second contrastive term aligning users with the images they have posted on Reddit. For users who 
have ever posted images, we extract CLIP embeddings of their images, forming user-image pairs $(u, \text{img}_u)$ and  similarly, the user-image loss is:

\vspace{-1em}
\[
\mathcal{L}_{u,\text{img}} = \frac{1}{B'} \sum_{j=1}^{B'} 
    \mathrm{CE}(\mathbf{z}_{u_j}^\top [\text{img}^t_{t_1}, \ldots, \text{img}^t_{t_{B'}}], 
        j 
    \bigr)
\]
\label{eq:loss-ut}
\vspace{-0.8em}

where $B' \leq B$ is the subset of users with available image-post text. The total 
loss combines both objectives: $\mathcal{L} = \mathcal{L}_{u,s} + \lambda \mathcal{L}_{u,\text{img}}$, 
with $\lambda = 1.0$ weighting the image contrastive loss. This dual supervision ensures 
learned embeddings capture not only where users engage (subreddit structure) but also 
what visual content they share, which we hypothesize as a critical proxy for downstream persona-driven image 
personalization (\cref{sec:exp_results}). Other training details, ablations and hyperparameter settings are discussed in \cref{sec:supp-gnn}

\vspace{-1em}\paragraph{Persona Verbalization} 
To construct a user's natural language persona from their subgraph, we utilize the learned attention weights $\alpha_{us}$ from \cref{eq:GATConv2} as indicators of importance. These weights represent how strongly the encoder associates each subreddit $s$ with user $u$ when modeling their preferences. For a given user, we aggregate attention values across their neighborhood and normalize them to obtain:
\(
\tilde{\alpha}_{us} = \frac{\alpha_{us}}{\sum_{s' \in \mathcal{N}(u)} \alpha_{us'}}
\)
This yields a probability distribution over subreddits that captures the user's attention-weighted engagement profile, highlighting which communities most characterize their interests and behaviors.

We allocate a fixed token budget $B=4{,}096$ across the user's 
top subreddits proportional to $\tilde{\alpha}_{us}$. For each subreddit $s$ 
receiving $B_s = \lfloor B \cdot \tilde{\alpha}_{us} \rfloor$ tokens, we sample the 
user's highest-engagement posts and comments (ranked by score, deduplicated, filtered 
for removed/deleted content), extracting representative text that reveals their interests, 
communication style, and aesthetic preferences. This attention-guided sampling ensures 
personas reflect what the graph encoder learned as behaviorally salient, rather than 
arbitrarily frequent interactions.

The sampled content is then presented to an MLLM with a structured prompt 
requesting a concise, coherent persona description that captures the user's visual 
interests and aesthetic tendencies. For example, a user active in r/analog, 
r/AnalogCommunity, and r/itookapicture with high attention weights and comments praising 
"grain texture" and "natural light" might yield: \textit{"Film photography enthusiast 
favoring high-grain black-and-white aesthetics with natural lighting and vintage 
tonality."}\cref{sec:supp-persona}.

\vspace{-0.5em}\subsection{Image Personalization from Personas}

We use the resulting natural language persona to guide zero-shot image personalization via persona-conditioned prompt rewriting. Instead of directly editing the prompt content, we use the persona as a \textit{roleplay instruction} that defines how the model(GPT-4o) should interpret and express visual preferences. Specifically, we prepend the persona as a first-person conditioning statement (e.g., \textit{"You are a film photography enthusiast favoring high-grain black-and-white aesthetics..."}), allowing the model to personalize generation from its own perspective. This enables implicit modulation of stylistic, compositional, and semantic attributes in line with the user's aesthetic tendencies. We detail the prompting format, decoding strategies, and evaluation protocols in \cref{sec:supp-gen}.

\section{Experiments \& Results}
\label{sec:exp_results}
In this section, we systematically evaluate \zipp against state-of-the-art personalization methods. We demonstrate that natural language persona conditioning not only enables effective zero-shot alignment (\cref{tab:main_results}) but also solves critical shortcomings in existing methods, namely the collapse of intra-user diversity (pluralism) (\cref{ssec:pluralism}) and the bias towards average-user demographics (alignment) (\cref{ssec:generalization}). We validate our approach across four datasets for zero-shot, few-shot, and finetuning paradigms (\cref{tab:main_results}). We also analyze the data efficacy of these methods (\cref{fig:shot-lineplot}) and corroborate our automated metrics with a comprehensive human user study (\cref{ssec:user_study}).

\subsection{Experimental Setup}
\paragraph{Benchmarks.}
\label{ssec:benchmarks}
We evaluate our method across four complementary datasets which assess personalization using preferences and generation history. For preferences, we utilize Flux-2-pro\_t2i\_human\_preference\footnote{\url{https://huggingface.co/Rapidata/datasets}} from \textbf{RapidData}, which provides explicit preference pairs alongside detailed user demographics (language, country, profession, gender, and age). \textbf{MovieLens}~(ML) is a movie-watching dataset covering 610 users, each with 20 to $\sim$2{,}600 interactions. Following~\cite{kim2025draw}, we convert each movie\'s metadata (title, genre, keywords, and description) into text prompts via the TMDB API; consistent with this prior work, we rely exclusively on these prompts for evaluation, as the dataset contains no images directly tied to T2I model outputs. For historical prompts, we evaluate on \textbf{\zipbench} (our proposed benchmark of 1.5K users, natural-language personas, and 40K generated images mined via our approach) and \textbf{PIP}~\cite{chen2024tailored}, which contain 300k extensive historical user generations from 3.1K users.

\paragraph{Baselines.}
\label{ssec:baselines}
We compare against five state-of-the-art personalization methods:
Random Baseline \textbf{LLM} (prompt rewriting without any context)
\textbf{TV}~\cite{chen2024tailored} (prompt rewriting via retrieval over $\leq$3 historical prompts),
\textbf{DrUM}~\cite{kim2025draw} (coreset sampling + per-user adapter training on 100+ prompts),
\textbf{ViPer}~\cite{salehi2024vipervisualpersonalizationgenerative} (trained VLM for preference extraction from user feedback)
We compare these methods against our Zero-Shot, Few-Shot persona conditioning.

\paragraph{Metrics.}
\label{ssec:metrics}
Following prior work~\cite{chen2024tailored,kim2025draw,von2024fabric}, we adopt \textit{ClipScore} (CLIP cosine similarity between generated image and ground-truth prompt)  as our primary automated metric for \textbf{personalization}, following the exact setup of ~\cite{kim2025draw}.
However, recent work has shown that CLIP-based metrics miss nuanced stylistic and compositional preferences that align with human evaluations ~\cite{lee2025personalized}. We therefore additionally report \textit{PIGReward}, a preference reward metric that evaluates image alignment through multi-dimensional assessment of style, mood, subject fidelity, and cultural coherence.
For \textbf{demographic alignment}, we report both the standard average and a weighted average that corrects for over- and under-represented demographic groups. Weights are computed via Iterative Proportional Fitting (IPF) applied to the population marginals for country, age, and gender sourced from the UN World Population Prospects 2024\footnote{\url{https://population.un.org/wpp/}}.
To measure \textbf{pluralism} of personalized outputs, we use the \textit{Clip Maximum Mean Discrepancy (CMMD)}~\cite{jayasumana2024rethinkingfidbetterevaluation} to measure distributional alignment between reference and personalized images. CMMD is an unbiased alternative for FID, and is more human aligned.

\subsection{Personalization from Natural Language Persona Conditioning}
\label{ssec:zero_shot}

We first establish that natural-language personas serve as an effective and universal personalization signal that bring consistent improvements in zero-shot and few-shot settings.
\Cref{tab:main_results} reports the unified evaluation across all four benchmarks and all baselines.

\subsubsection{\zipbench}
\begin{figure*}[t]
\centering

\begin{subfigure}[t]{0.63\linewidth}
    \centering
    \includegraphics[width=\linewidth]{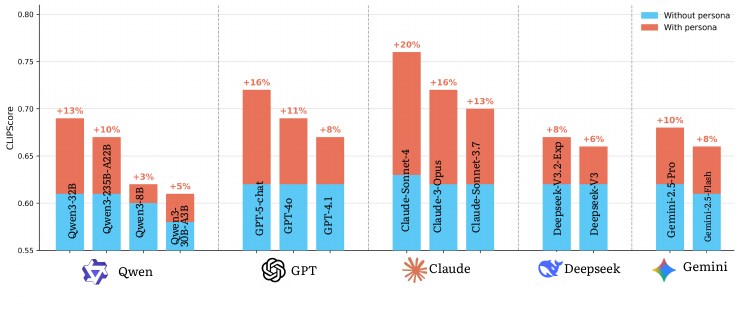}
    \label{fig:imagealign_by_family}
\end{subfigure}
\hfill
\begin{subfigure}[t]{0.36\linewidth}
    \centering
    \includegraphics[width=\linewidth]{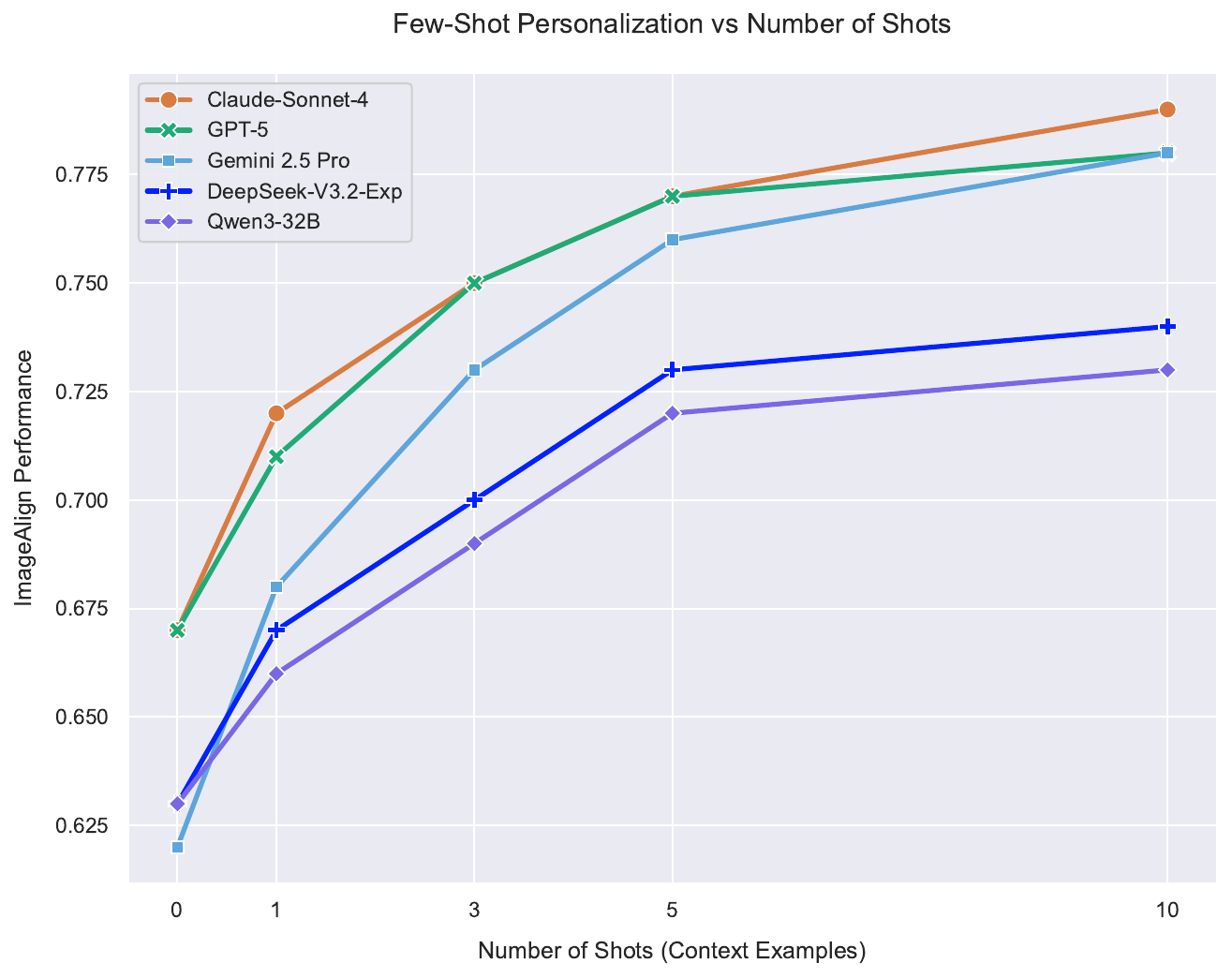}
    \label{fig:shot-lineplot}
\end{subfigure}

\caption{
\textbf{Effect of persona conditioning on generation quality.}
(a) ClipScores by model family with persona conditioning. Stacked bars show baseline (blue) and improvement from persona conditioning (red). All models benefit from persona conditioning ranging from 3\% (Qwen3-30B-A3B) to 20\% (Claude-Sonnet-4). Frontier models achieve the strongest gains (13--20\%), outperforming open-source alternatives (3--13\%).
(b) Performance increase as the number of in-context persona examples increases.
}
\label{fig:persona_conditioning_results}

\end{figure*}

\cref{fig:imagealign_by_family} shows the \zipbench evaluation for ClipScore for 14 LLMs across five model families (Qwen, GPT, Claude, DeepSeek, Gemini), with and without persona conditioning (red and blue).
Persona conditioning yields consistent improvements across \emph{all} (frontier and open-source) evaluated models.
Frontier models outperform open-source models due to their stronger instruction-following and persona interpretation capabilities, corroborated by recent roleplay benchmarks~\cite{du2025twinvoice} where Claude ranks highest.
Mixture-of-experts models (Qwen3-30B-A3B, Qwen3-235B-A22B) surprisingly underperform their dense counterparts despite larger parameter counts; on manual analysis we found that these models are over-personalizing. Performance of all baselines on \zipbench is given in \Cref{tab:main_results}. Extended results across remaining metrics and hyperparameters are reported in
\cref{tab:zip_clip_imagealign_table} and \cref{sec:supp-results}

\subsubsection{Public Benchmarks.} \Cref{tab:main_results} reports CLIPScore and PIGReward across \zipbench, PIP, RapidData, and MovieLens. For fair comparison, we use GPT-4o as the underlying LLM for all prompt-rewriting methods.
Even in the \textbf{zero-shot} setting, \MyModel~already surpasses the fine-tuned DrUM baseline on three of four benchmarks, demonstrating that a compact natural-language persona captures preference signals as effectively as dense interaction histories and per-user adapter training.
In the \textbf{few-shot} regime, \MyModel~(5-shot) achieves the highest scores across all four benchmarks on both metrics. At the same shot count, \MyModel~(3-shot) consistently outperforms TV~(3-shot), confirming that persona priors provide a stronger conditioning signal than retrieval over raw prompt histories alone.
The only benchmark where DrUM remains competitive is PIP, which we attribute to its adapter being trained on the same dense prompt histories that define PIP users; on the remaining benchmarks, persona conditioning surpasses per-user fine-tuning without any gradient updates.
To adapt our method to external benchmarks, we construct PIP user personas by embedding each user's prompt history via a CLIP text encoder and retrieving nearest-neighbor personas from our mined set (cosine similarity $\geq$0.7, covering 89\% of PIP users); for RapidData and MovieLens, where only demographic variables and profession are available, we condition on these attributes directly.
As a result, the strongest gains appear on \zipbench, where full natural-language personas are available, followed by RapidData and MovieLens (demographics only), and finally PIP (retrieved personas). To investigate the relationship between persona richness and personalization capability in detail, we ablate summarized profiles across different token budgets in~\Cref{fig:token-budget}.

\subsubsection{Pluralistic Preference Alignment}
\label{ssec:pluralism}
Effective personalization must capture not only \emph{what} a user prefers on average, but also \emph{how} their preferences vary across different contexts. A person who prefers `muted' and `animated' portraits could prefer `bright' street photography should not get similar output for every prompt.
We formalize this as \textit{pluralistic alignment}: a personalization method should achieve high average personalization \emph{while preserving} the distributional spread of a user's true visual preference space. To measure the distributional alignment of target and generated images, we employ the CMMD metric, which is essentially the squared MMD of target and generated image's clip embeddings, using with the Gaussian RBF kernel, first introduced by ~\cite{jayasumana2024rethinkingfidbetterevaluation}.
Lower MMD indicates closer alignment between the generated and ground-truth image distributions. We report these in \Cref{tab:main_results},\somesh{Qual Analysis}.
ViPer assigns the same style attributes across all contexts for the same user, making it the most susceptible to pluralistic misalignment---it achieves the worst CMMD (0.55).
DrUM exhibits moderate CMMD (0.31): while its per-user adapter captures individual preferences, its coreset sampling anchors generation around a narrow slice of the user's history.
TV (0.25) benefits from retrieval diversity but still collapses when the prompt pool is small.
In contrast, \MyModel achieves the \emph{lowest} CMMD (0.16 at 5-shot) while maintaining the \emph{highest} alignment scores, demonstrating that persona conditioning preserves intra-user preference diversity.
This occurs because personas encode multi-faceted preference priors---spanning aesthetics, topics, culture, and affect---that modulate generation contextually, rather than anchoring on a fixed reference set.

\subsubsection{Demographic Generalization}
\label{ssec:generalization}
Survey-based optimization and evaluation have historically suffered from sampling bias toward WEIRD (Western, Educated, Industrialized, Rich, Democratic) populations~\cite{henrich2010weirdest}.
Henrich et al.\ showed that up to 96\% of behavioral science participants come from WEIRD societies---representing only ${\sim}$12\% of the global population---yet findings are routinely generalized as universal.
This bias extends to AI systems: Santurkar et al.~\cite{santurkar2023whose} demonstrated that language models disproportionately reflect the opinions of specific U.S.\ demographic groups, with alignment varying systematically across age, education, and political affiliation.
To mitigate such biases, survey organizations like Pew Research apply Iterative Proportional Fitting (IPF, also known as ``raking'') to reweight responses so that reported aggregates reflect true population distributions rather than the convenience sample~\cite{santurkar2023whose}.
We find that these same biases have crept into image personalization and preference alignment benchmarks.
Critically, most existing benchmarks---including PIP~\cite{chen2024tailored}, HPD~\cite{wu2023humanpreferencescorev2}, and Pick-a-Pic~\cite{kirstain2023pick}---do not report individual or aggregate demographic distributions at any level, making it impossible to analyze or alleviate these biases.
To address this gap, we utilize RapidData's T2I preference datasets, a large-scale image preference dataset where each preference pair is explicitly tagged with anonymized group-level demographics (language, country, profession, age, gender).
RapidData serves as a demographically transparent alternative to HPD and Pick-a-Pic, enabling the first systematic analysis of demographic bias in image personalization.
\paragraph{IPF normalization.}
Raw aggregate scores can mask systematic disparities.
We extract the true demographic marginal distributions using UN World Population Prospects 2024, which provides population-level statistics for each demographic axis.
We then apply Iterative Proportional Fitting (IPF)~\cite{deming1940least} to reweight per-subgroup scores, ensuring that the reported aggregate reflects a balanced population rather than the majority-skewed sample.
We emphasize the importance of annotating preference datasets with anonymized demographics to ensure both privacy and equitable alignment evaluation. We show results for 
 stratified ClipScore and PIGReward by demographic subgroup in \Cref{sec:supp-generalization}.
Since the evaluations themselves are skewed, optimization methods that maximize aggregate population scores are most susceptible to this bias.
DrUM shows the \emph{highest} drop in scores after IPF normalization , as its adapter training absorbs the distributional biases of the majority-skewed training data.
In contrast, \MyModel shows the highest alignment in all subgroups because the LLM takes the persona into its generation context.
Even in the zero-shot setting, \MyModel achieves better or comparable performance to trained or few-shot models.
ViPer shows very little relative degradation after normalization ; we attribute this to the fact that ViPer's visual preference representations are trained on synthetic visual personas that are less susceptible to demographic skew in interaction histories.
On subgroup-level analysis, we observe clear stereotypical trends: DrUM shows poor alignment with Older (Age 50+), African, Russian, and South American populations and individuals in manual trades.
We also observe that frontier models (Claude, GPT, Gemini) show far better demographic alignment compared to open-source alternatives (Qwen, DeepSeek), consistent with the broader instruction-following gap observed in~\cref{ssec:zero_shot}.
A detailed visualization of per-subgroup score distributions and IPF-normalized aggregates is provided in~\cref{sec:supp-generalization}.

\somesh{1. For Viper it will use the exact same style/theme for the common user across all prompts Street, Cat, Girl 2. For DrUM it will bias the Brazil to European style 3. Only ours (0 and few) will show correct emotions for all faces, presence of people on street 4. Ours 0-shot will miss simple things but that can only be inferred from few shot examples like specific art styles 4. TV has low generalization so it will miss some style attributes}

\begin{table*}[t]
\centering
\begin{adjustbox}{width=\linewidth}
\setlength{\tabcolsep}{6pt}
\renewcommand{\arraystretch}{1.15}

\begin{tabular}{cl
cc
cc
!{\vrule width 0.6pt}
cc
cc
!{\vrule width 1.2pt}
cc
c}

\toprule
\textbf{Train} & \textbf{Method}

& \multicolumn{2}{c}{\textbf{\zipbench}}
& \multicolumn{2}{c}{\textbf{PIP}}

& \multicolumn{2}{c}{\textbf{MovieLens}}
& \multicolumn{2}{c}{\textbf{RapidData}}

& \multicolumn{2}{c}{\textbf{Demographic}}
& \textbf{Pluralism} \\

\cmidrule(lr){3-4}
\cmidrule(lr){5-6}
\cmidrule(lr){7-8}
\cmidrule(lr){9-10}
\cmidrule(lr){11-12}

& 
& \textbf{CS$\uparrow$} & \textbf{PIG$\uparrow$}
& \textbf{CS$\uparrow$} & \textbf{PIG$\uparrow$}
& \textbf{CS$\uparrow$} & \textbf{PIG$\uparrow$}
& \textbf{CS$\uparrow$} & \textbf{PIG$\uparrow$}
& \textbf{CS$\uparrow$} & \textbf{PIG$\uparrow$}
& \textbf{CMMD$\downarrow$} \\

\specialrule{0.5pt}{0pt}{0pt}

\multirow{2}{*}{\makecell{zero\\shot}}

& GPT-4o
& 59.1 & 65.3 & 57.7 & 61.4
& 73.7 & 58.5 & 67.3 & 56.0
& 66.9 & 55.1 & 0.49 \\

& \MyModel
& 61.8 & 73.4 & 60.2 & 67.3
& 76.4 & 64.2 & 73.9 & 61.5
& 72.8 & 60.9 & 0.42 \\

\specialrule{0.5pt}{0pt}{0pt}

\multirow{4}{*}{\makecell{few\\shot}}

& FABRIC (1)
& --  & -- & 63.7 & --
& 75.1 & -- & -- & --
& - & - & - \\

& TV (3)
& 62.4 & 72 & 63.1 & 66.5
& 80.1 & 68.7 & 77.4 & 65.3
& 73.8 & 60.1 & 0.25 \\

& \MyModel (3)
& \valgood{66.7} & \valgood{77.5} & 65.2 & \valgood{69.9}
& \valgood{80.4} & \valgood{71.4} & \valgood{77.9} & \valgood{68.2}
& \valgood{76.8} & \valgood{67} & \valgood{0.19} \\

& \MyModel (5)
& \valbest{68.5} & \valbest{82.8} & \valgood{66.3} & \valbest{72.1}
& \valbest{80.9} & \valbest{77.9} & \valbest{78.3} & \valbest{74.0}
& \valbest{77.4} & \valbest{73.0} & \valbest{0.16} \\

\specialrule{0.5pt}{0pt}{0pt}

\multirow{2}{*}{\makecell{fine\\tune}}

& DrUM
& 65.6 & 74.4 & \valbest{66.9} & 68.0
& 72.8 & 66.2  & 70.3 & 63.5
& 64.7 & 56.8 & 0.31 \\

& ViPer
& -- & -- & -- & --
& 76.1 & 70.5 & 74.1 & 63.1
& 73.7 & 60.8 & 0.55 \\

\bottomrule
\end{tabular}

\end{adjustbox}
\caption{Pluralistic \& Demographic alignment.\harini{add a row over the first row }\textbf{Unified evaluation: cross-benchmark, demographic, and pluralistic alignment.}
CS~=~CLIPScore, PIG~=~PIGReward, CMMD~=~CLIP Maximum Mean Discrepancy ($\downarrow$~is~better).
We report personalization benchmarks and methods for (i) personalization from historical prompts (\zipbench, PIP) (ii) from preference pairs (MovieLens, RapidData) and across 0-shot, few-shot (FABRIC,TV), and fine-tuning baselines (DrUM, ViPer). \MyModel achieves the highest scores across all benchmarks and the lowest CMMD, demonstrating simultaneous personalization quality, demographic equity, and pluralistic diversity preservation.}
\label{tab:main_results}
\end{table*}

\section{User Study}
\label{ssec:user_study}
To complement our automated metrics, we conduct a controlled human evaluation assessing pairwise preference of different personalization methods. We recruited 50 diverse participants from a large institution, we collect their demographics, professions, and hobbies to create the natural-language persona. Participants performed pairwise A/B testing over 20 different prompts, selecting the image that better reflected their visual preferences compared across methods. This results into 1000 annotations, in the \textbf{persona versus no-persona} condition, our zero-shot \MyModel achieved a \textbf{79\% win rate}, demonstrating a strong, recognizable preference for persona-aligned outputs over generic generations. When testing personalized methods against one another, \MyModel (both 0-shot and 3-shot) outperformed all established baselines:
\begin{itemize}
    \item \textbf{\MyModel (0-shot) vs. TV (3-shot):} 56\% win rate in favor of \MyModel.
    \item \textbf{\MyModel (3-shot) vs. DrUM (Fine-tuned):} 58\% win rate, despite DrUM requiring per-user adapter training on over 100 historical prompts.
    \item \textbf{\MyModel (3-shot) vs. ViPer (Fine-tuned):} 65\% win rate
\end{itemize}
These results reaffirm our quantitative findings: persona-guided conditioning achieves superior, highly pluralistic personalization that aligns closely with human judgment.

\section{Conclusion and Impact}
\vspace{-0.5em}
We presented \textbf{ZIPP}, a zero-shot framework for image personalization through natural language personas. By grounding user preferences in interpretable textual descriptions, ZIPP personalizes diffusion outputs without fine-tuning or user data. Experiments and human evaluations demonstrate that persona conditioning enhances visual alignment, and that mined personas are accurate, coherent, and preferred by users. We also propose methods to evaluate and optimize the pluralism and demographic aligned of existing methods and benchmarks. All datasets are derived solely from publicly available Reddit content, with rigorous anonymization and no user re-identification. We further discuss data handling, consent considerations, and the broader ethical implications of persona-based modeling in the Appendix.

\bibliography{main}
\bibliographystyle{iclr2026_conference}

\newpage
\appendix
{\par\centering\Large\textbf{Appendix}\par}

\startcontents[appendix]
\printcontents[appendix]{}{1}{}

\clearpage

\section{Qualitative Examples}
\label{sec:supp-qualitative}

\subsection{Comparison with other methods}
\begin{figure}
    \centering
    \includegraphics[width=0.8\linewidth]{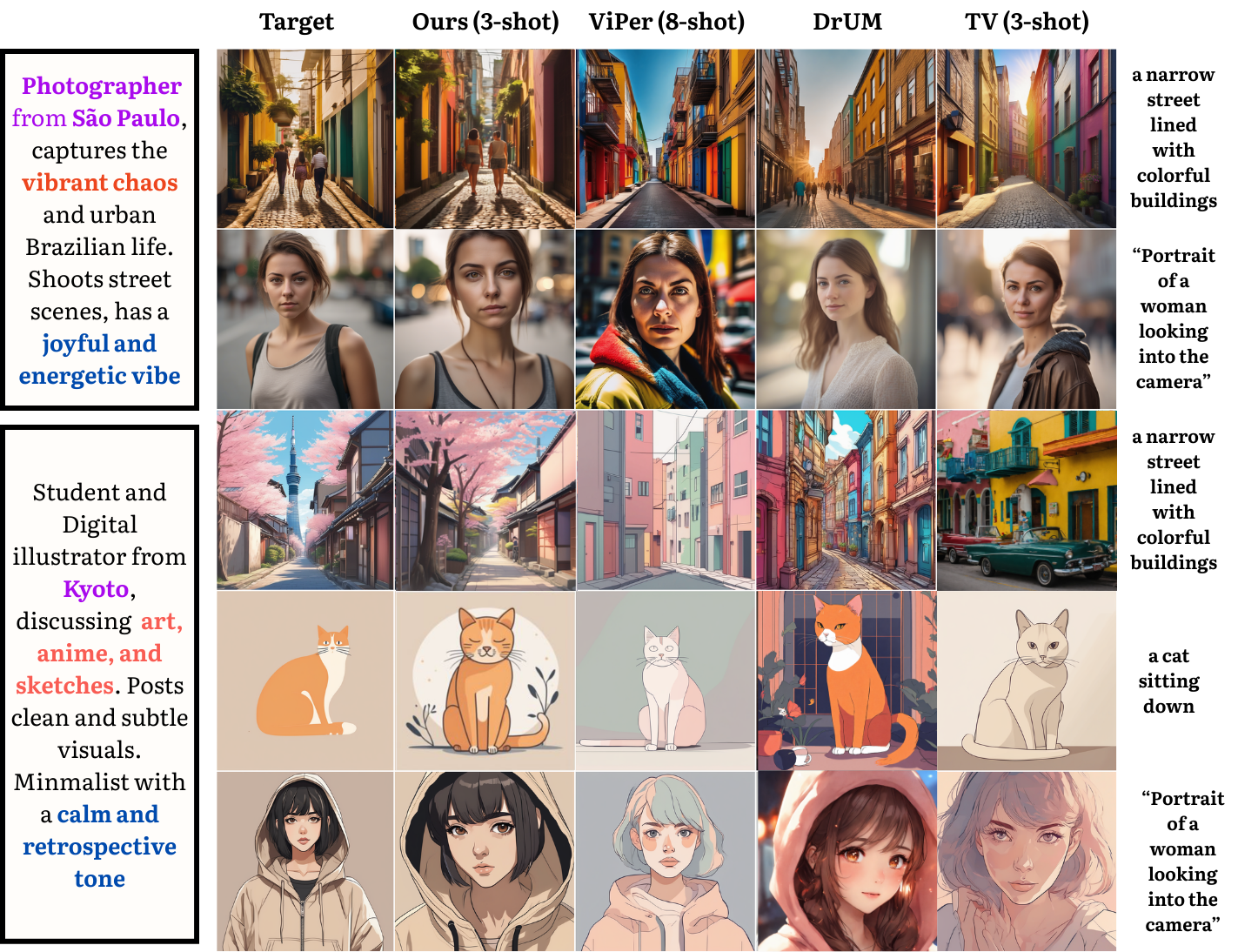}
    \caption{Qualitative Comparison \somesh{complete caption}}
    \label{fig:qual-example}
\end{figure}

\somesh{Quoted Prompt rewriting examples}
\somesh{section and subsection intro}

\subsection{Persona: Reddit User Activity} 
\begin{figure}[ht]
    \centering
    \includegraphics[width=0.95\linewidth]{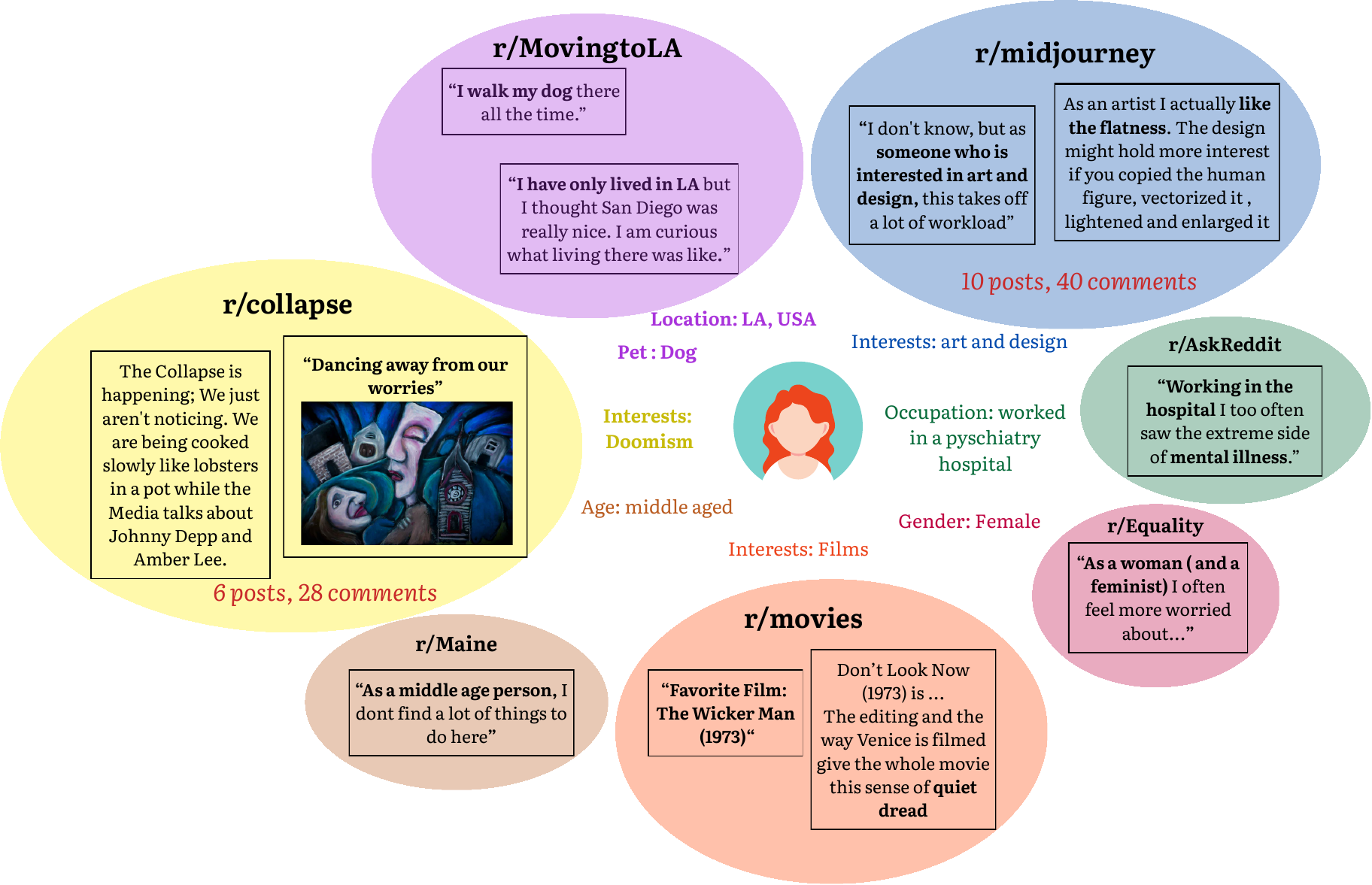}
    \caption{The figure shows how a coherent persona can be inferred by aggregating signals from a user’s posts and comments across different Reddit communities.}
    \label{fig:reddit_history}
\end{figure}

\cref{fig:reddit_history} shows how coherent personas can be inferred by aggregating signals from user activity across diverse Reddit communities. This user's posts and comments span communities ranging from r/MovingtoLA and r/midjourney to r/collapse, r/movies, and r/Equality. By analyzing the content and themes of her engagement, we infer a middle-aged woman living in Los Angeles who worked in a psychiatric hospital setting. Her r/collapse participation reveals preoccupation with societal breakdown and apocalyptic themes—exemplified by her comment ``The Collapse is happening; We just aren't noticing. We are being cooked slowly like lobsters in a pot while the Media talks about Johnny Depp and Amber Lee.'' Her r/AskReddit activity surfaces personal history (``Working in the hospital I too often saw the extreme side of mental illness''), while r/Equality posts demonstrate feminist perspectives and social justice concerns. Geographic roots emerge from r/Maine discussions referencing coastal upbringing, and r/midjourney activity confirms active engagement with AI art generation.

We extract these posts and comments and convert them into a structured natural language profile using the verbalization algorithm described in \cref{alg:persona_verbalization}. Listing \ref{tab:persona_examples} shows the resulting profile, which accurately prioritizes the most persona-revealing content.

However, these structured profiles cannot be directly used by LLMs for roleplay, as shown in prior work \cite{chen2025persona}. We therefore convert them into a roleplay-compatible system prompt (see Listing \ref{tab:persona_extraction_template}). The final system prompt is shown in Listing \ref{tab:persona_rewrite_template}.

\somesh{highlight figure and text with same colors}

Below is the user's persona created from their online activity.
\phantomsection\label{tab:persona_examples}
\begin{quote}
\textit{You are a woman in your 30s–40s living in Los Angeles, originally from coastal Maine. You work in a hospital psychiatric setting and also explore art in your free time. You spend time online creating and discussing ideas about art, technology, science, and social issues. Your work explores imaginative and speculative themes, often touching on doomism, strange futures, and cultural commentary. Growing up along Maine’s foggy coast, pine forests, cliffs, and small towns shaped your sensibility. You are drawn to quiet, atmospheric environments and often reference landscapes, weather, and coastal life. You are curious about the strange and unexplained. Paranormal stories, folklore, and unusual phenomena interest you, though you often approach them with curiosity rather than certainty. You have a pet dog.}
\end{quote}


\somesh{Text deepdive}

\subsection{Pluralistic Visual Prefences}
\begin{figure}[ht]
    \centering
    \includegraphics[width=0.95\linewidth]{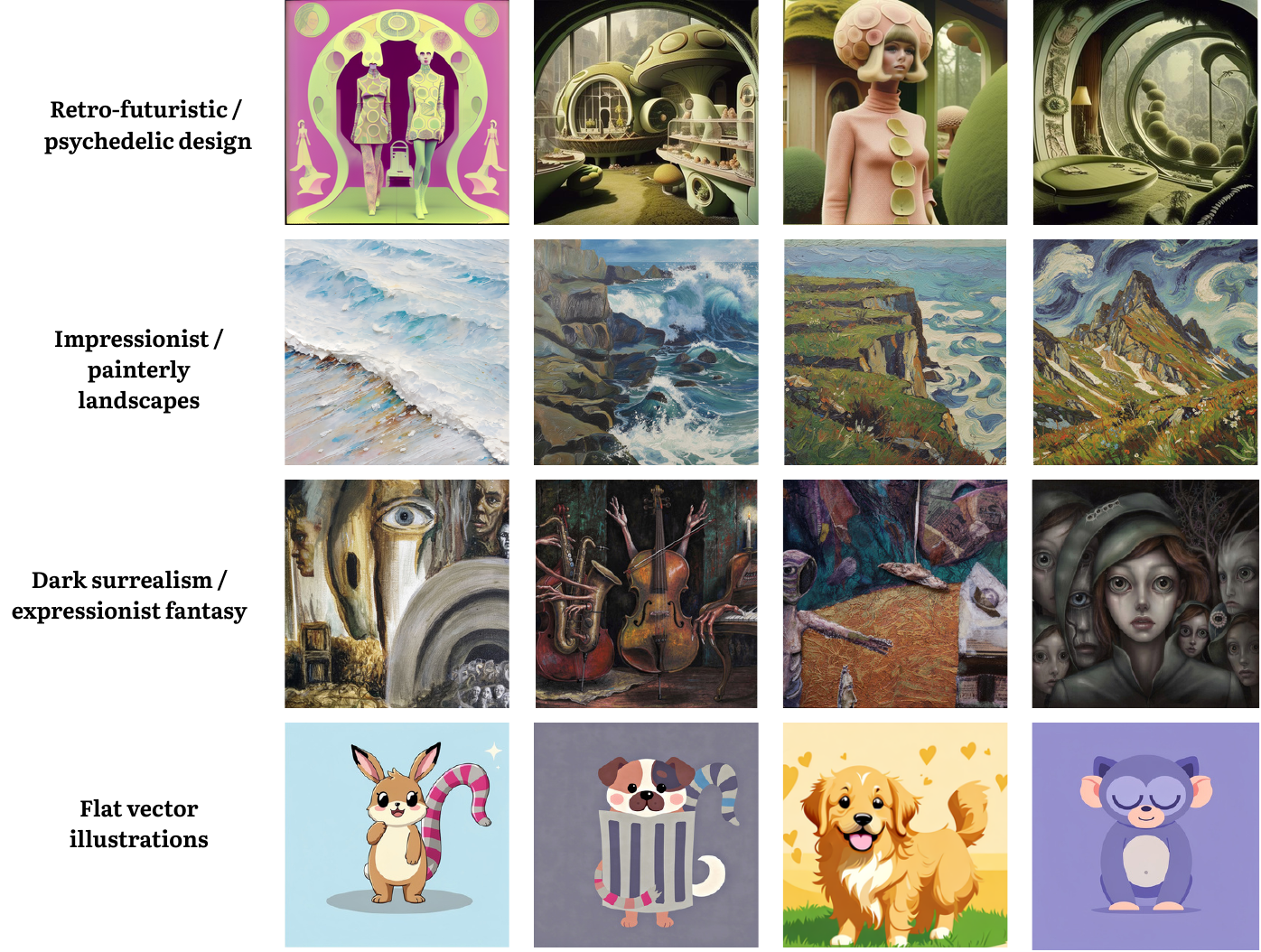}
    \caption{The figure shows this user's pluralistic preferences.}
    \label{fig:pluralism}
\end{figure}

This user's visual preferences reveal pluralistic aesthetics that directly map to persona elements (\cref{fig:pluralism}). The \textbf{dark surrealist fantasy} style—featuring expressionist figures, ominous atmospheres, and existential symbolism—directly reflects her doomism and collapse-oriented worldview, translating apocalyptic anxiety into visual metaphor. The \textbf{impressionist painterly landscapes} depicting windswept coasts, rocky shores, and atmospheric seascapes echo her Maine coastal upbringing, where ``foggy coast, pine forests, cliffs, and small towns shaped [her] sensibility.'' The \textbf{retro-futuristic psychedelic designs} align with her speculative and imaginative art interests, while \textbf{flat vector illustrations} of companion animals reflect her mentioned pet dog. This demonstrates how verbal personas encode not just demographic facts, but deeper psychological and aesthetic tendencies that manifest consistently across creative contexts.

\section{Implementation details}
\label{sec:impl_details}

\subsection{Graph encoder and objectives}
\label{sec:supp-gnn}
We provide full definitions for edge features, encoder layers, and training losses. Edge attributes include degree-corrected weights $w_{us}=\log(1+t_{us})/\sqrt{c_uc_s+\varepsilon}$ and comment fractions $f^{c}_{us}=c_{us}/\max(t_{us},1)$. Attention uses $e_{us}=[w_{us},f^{c}_{us}]$ inside GATv2 with $K=4$ heads; we stack two layers with residuals and L2-normalize. The link-prediction loss $\mathcal{L}_{\mathrm{link}}$ is binary cross-entropy with negative sampling. The contrastive terms $\mathcal{L}_{uv}$ and $\mathcal{L}_{ui}$ use temperature-scaled InfoNCE with in-batch negatives and L2-normalized heads. We also consider using unnormalized weights however due to high node degrees as seen in \cref{sec:supp-degree} the user embeddings converged towards larger subreddits (e.g. r/AskReddit).

\subsection{Persona verbalization}
\label{sec:supp-persona}
We describe the attention aggregation $\bar{\alpha}_{C,s}$ computation, normalization $\tilde{\alpha}_{C,s}$, token/character budget $B$ allocation across subreddits, and selection of representative posts/comments. We also include cluster-level TF–IDF and NER extraction, weighted by $\tilde{\alpha}_{C,s}$, and discuss ties with our earlier TF–IDF-based script. For all of our experiments we use a token budget of 4096, and limit to 50 subreddits, we use GPT-4o for all verbalization for Personas, Open source LLMs like Qwen often struggled with Social Media jargon. The algorithm is detailed in \cref{alg:persona_verbalization}

\subsection{Prompt rewriting}
\label{sec:supp-gen}
We have given the prompt templates for each setting used in the paper. We use a temperature of 0.7 for every LLM. All other decoding parameters follow the model defaults.

\subsection{Diffusion model settings}
\label{sec:supp-diffusion}
We evaluate ZIPP using several widely adopted text-to-image diffusion models across both open-source and proprietary families. All models are used in inference-only mode without any fine-tuning.

\paragraph{Models}
We use the following diffusion backbones:
\begin{itemize}
    \item Stable Diffusion XL (SDXL)
    \item Stable Diffusion 3.5
    \item Stable Diffusion 3
    \item Flux.1-dev
\end{itemize}

Unless explicitly indicated, all reported scores use SDXL as the default model.

\paragraph{Hyperparameter settings}
To ensure consistent comparisons across users, personas, and model families, we fix the following generation settings:
\begin{itemize}
    \item \textbf{Inference steps:} 50 steps for all models.
    \item \textbf{Guidance scales:} We use the default guidance scales for each of these models
    \begin{itemize}
        \item SDXL: 7.0
        \item SD3.5: 5.0
        \item SD3: 5.0
        \item Flux: 4.0
    \end{itemize}
    \item \textbf{Randomness:} Three random seeds are used for each prompt. All metrics are computed over these samples.
\end{itemize}

We do not add negative prompts or handcrafted style keywords to ensure the evaluation remains consistent with the baselines. This allows us to isolate and measure the impact of persona conditioning itself.

\paragraph{Batching and Hardware}
All images are generated using 8\,\texttimes\,A100 80GB GPUs with fp16 inference.  

\subsection{IPF}
Standard aggregate metrics weight each user equally, implicitly reflecting the demographic composition of the evaluation sample.
When that sample is skewed as is common in crowdsourced preference datasets that over-represent WEIRD populations~\cite{henrich2010weirdest} raw aggregates can overstate performance on majority subgroups and mask failures on underrepresented ones.
Iterative Proportional Fitting (IPF, also known as \textit{raking})~\cite{deming1940least} is a classical post-stratification technique widely used in survey methodology~\cite{pewresearch_center,santurkar2023whose} to correct for such imbalances.

\paragraph{Problem setup.}
Let $\{u_i\}_{i=1}^N$ denote the evaluation sample, where each user is annotated with categorical demographic labels $d_k(u_i)$ across $K$ axes (e.g.\ age, gender, country, language, profession).
Let $s_i$ denote the per-user evaluation score (CLIPScore or PIGReward).
We are given census-level target marginals $p^*_k(c)$ specifying the desired population proportion of each category $c$ along axis $k$.
The goal is to compute a set of per-user weights $\{w_i\}$ such that the weighted marginal distribution along \emph{every} demographic axis matches the target, and then report the reweighted aggregate $\bar{s}_{\text{IPF}} = \sum_i w_i \cdot s_i$.

\paragraph{Algorithm.}
IPF operates by iteratively cycling through each demographic axis and rescaling user weights so that the weighted sample marginal for that axis matches the target (\Cref{alg:ipf_reweighting}).
Concretely, at each iteration the algorithm visits axis $k$ and, for every category $c$ in that axis, computes the current weighted proportion $\hat{p}_k(c) = \sum_{i:\,d_k(u_i)=c} w_i$.
It then multiplies the weight of every user belonging to category $c$ by the raking factor $f_k(c) = p^*_k(c) / \hat{p}_k(c)$, which up-weights underrepresented subgroups and down-weights overrepresented ones.
After adjusting all categories within an axis, weights are re-normalized to sum to one.
This process repeats across all $K$ axes until the maximum discrepancy between any weighted marginal and its target falls below a convergence threshold $\epsilon$.
Under mild non-degeneracy conditions (no structural zeros in the contingency table), IPF is guaranteed to converge to the unique set of weights satisfying all marginal constraints simultaneously~\cite{deming1940least}.

\begin{algorithm}[t]
\small
\caption{IPF-Normalized Evaluation}
\label{alg:ipf_reweighting}
\begin{algorithmic}[1]
\Require Sample users $\{u_i\}_{i=1}^{N}$, demographic axes $\mathcal{K} = \{1,\dots,K\}$, per-user labels $d_k(u_i)$, per-user scores $s_i$, census target marginals $\{p^*_k(c)\}$, threshold $\epsilon$.
\Ensure IPF-reweighted aggregate score $\bar{s}_{\text{IPF}}$.
\State Initialize weights: $w_i \leftarrow 1/N$ for all $i \in \{1,\dots,N\}$
\Repeat
    \For{each demographic axis $k \in \mathcal{K}$}
        \For{each category $c$ in axis $k$}
            \State Compute weighted sample marginal: $\hat{p}_k(c) \leftarrow \sum_{i:\, d_k(u_i) = c} w_i$
            \State Compute raking factor: $f_k(c) \leftarrow p^*_k(c)\, /\, \hat{p}_k(c)$
            \State Adjust weights: $w_i \leftarrow w_i \cdot f_k(c)$ for all $i$ with $d_k(u_i) = c$
        \EndFor
        \State Re-normalize: $w_i \leftarrow w_i \,/\, \sum_{j=1}^{N} w_j$ \Comment{Ensure $\sum_i w_i = 1$}
    \EndFor
\Until{$\max_k \max_c \left| \hat{p}_k(c) - p^*_k(c) \right| < \epsilon$}
\State Compute reweighted score: $\bar{s}_{\text{IPF}} \leftarrow \sum_{i=1}^{N} w_i \cdot s_i$
\State \textbf{return} $\bar{s}_{\text{IPF}}$
\end{algorithmic}
\end{algorithm}

\paragraph{Interpretation.}
The resulting weights jointly satisfy all $K$ marginal constraints, ensuring the aggregate score reflects a demographically balanced population.
A method whose raw and IPF-reweighted scores are similar performs equitably across subgroups; a large gap signals that performance concentrates on the majority.
In our evaluation (\cref{ssec:generalization}), we apply IPF with $K=5$ axes (age, gender, country, language, profession) using marginals from the UN World Population Prospects 2024, and set $\epsilon = 10^{-4}$, which yields convergence within 5--8 iterations in practice.

\harini{add algo and figure}

\section{Algorithm}
\label{sec:supp-algorithm}

The detailed algorithm for learning user representations with a contrastive loss is provided in \cref{alg:user_rep_learning}, and the procedure for constructing a user’s natural language persona from the learned attention weights is described in \cref{alg:persona_verbalization}.

\begin{algorithm*}[ht]

\caption{Contrastive User Representation Learning}
\label{alg:user_rep_learning}
\begin{algorithmic}[1]
\Require Bipartite Graph $\mathcal{G}=(\mathcal{U}, \mathcal{S}, \mathcal{E})$, User Images $\mathcal{D}_{\text{img}} = \{(u, \mathbf{v}_{\text{img}})\}_{u \in \mathcal{U}'}$ where $\mathcal{U}' \subset \mathcal{U}$.
\State \textbf{Input Processing:}
\State Compute normalized edge weights: $w_{us} \leftarrow \frac{\log(1 + t_{us})}{\sqrt{d_u d_s + \epsilon}}$
\State Initialize $\mathbf{h}_u \sim \mathcal{N}(0, 0.01\mathbf{I})$ \Comment{Gaussian noise for users}
\State Initialize $\mathbf{h}_s \leftarrow \text{TextEncoder}(\text{Description}_s)$ \Comment{Pretrained embeddings from subreddit descriptions}
\State \textbf{Training Loop:}
\While{not converged}
    \State Sample batch $\mathcal{B}_u \sim \mathcal{U}$
    \State Sample one positive neighbor $s \in \mathcal{N}(u)$ uniformly per user \Comment{Balanced sampling}
    \State \textbf{Graph Encoding:}
    \State Apply GATv2 layers to update user representations:
    \State $\mathbf{h}_u \leftarrow \text{GATv2}(\mathbf{H}_{\mathcal{S}}, \mathcal{G}_{\mathcal{B}_u}, \mathbf{w})$ \Comment{See \cref{eq:GATConv2}}
    \State \textbf{Contrastive Optimization:}
    \State Compute User-Subreddit alignment: $\mathcal{L}_{\text{graph}} \leftarrow -\log \frac{\exp(\mathbf{h}_u \cdot \mathbf{h}_s / \tau)}{\sum_{k \in \mathcal{B}} \exp(\mathbf{h}_u \cdot \mathbf{h}_{s_k} / \tau)}$
    \State Compute User-Image alignment: $\mathcal{L}_{\text{img}} \leftarrow -\log \frac{\exp(\mathbf{h}_u \cdot \mathbf{v}_{\text{img}} / \tau)}{\sum_{k \in \mathcal{B}'} \exp(\mathbf{h}_u \cdot \mathbf{v}_{\text{img}_k} / \tau)}$
    \State \textbf{Update:}
    \State $\theta \leftarrow \theta - \eta \nabla (\mathcal{L}_{\text{graph}} + \lambda \mathcal{L}_{\text{img}})$
\EndWhile
\State \textbf{Output:} Learned user embeddings $\mathbf{h}_u$ and attention weights $\alpha_{us}$.
\end{algorithmic}
\label{alg:user-rep}
\end{algorithm*}

\begin{algorithm*}[ht]
\small
\caption{Attention-Guided Persona Verbalization}
\label{alg:persona_verbalization}
\begin{algorithmic}[1]
\Require Trained Graph Encoder, User set $\mathcal{U}$, Token Budget $B$, Subreddit posts $\mathcal{P}$.
\Ensure Natural language persona descriptions $\{\rho_u\}_{u \in \mathcal{U}}$.
\For{each user $u \in \mathcal{U}$}
    \State \textbf{Importance Inference:}
    \State Extract attention weights $\alpha_{us}$ from the trained GATv2 encoder (\cref{alg:user_rep_learning}).
    \State Compute importance score $s_{us} \leftarrow \alpha_{us}$ (or TF-IDF score as baseline).
    \State Normalize scores: $\tilde{s}_{us} \leftarrow s_{us} / \sum_{k \in \mathcal{N}(u)} s_{uk}$.
    \State Rank subreddits: $\mathcal{S}^* \leftarrow \text{argsort}_s(\tilde{s}_{us})$ descending.
    \State \textbf{Context Construction:}
    \State Initialize context buffer $C_u \leftarrow \emptyset$, current token count $T \leftarrow 0$.
    \For{subreddit $s \in \mathcal{S}^*$}
        \If{$T \ge B$} \State \textbf{break} \EndIf
        \State Determine slot budget $k_s$ proportional to $\tilde{s}_{us}$ \Comment{Rank decay prioritization}
        \State Retrieve posts $\mathcal{P}_s \leftarrow \text{GetHighEngagementPosts}(u, s)$
        \State Select diverse samples: $S_s \leftarrow \text{GreedySample}(\mathcal{P}_s, k_s)$ \Comment{Maximize diversity}
        \State Update context: $C_u \leftarrow C_u \cup S_s$
        \State Update count: $T \leftarrow T + \text{Length}(S_s)$
    \EndFor
    \State \textbf{Verbalization:}
    \State Construct prompt with template \Comment{See \cref{tab:persona_extraction_template}}
    \State Generate persona: $\rho_u \leftarrow \text{LLM}(\text{Prompt}, C_u)$
\EndFor
\State \textbf{return} $\{\rho_u\}_{u \in \mathcal{U}}$
\end{algorithmic}
\label{alg:attention}
\end{algorithm*}

\clearpage

\section{Prompts}
\label{sec:supp-prompts}

This section provides the complete prompt templates used throughout our experiments. We present five key templates: (1) \textbf{Persona Extraction} (\cref{tab:persona_extraction_template}), which transforms raw user activity into natural language personas; (2) \textbf{Zero-Shot Baseline} (\cref{tab:zs_prompt_template}), which rewrites prompts without persona conditioning to establish baseline performance; (3) \textbf{Zero-Shot Personalized} (\cref{tab:zs_personalized_prompt_template}), which incorporates persona conditioning for cold-start personalization; (4) \textbf{Few-Shot Personalized} (\cref{tab:fs_personalized_prompt_template}), which combines persona traits with example prompts for stronger alignment; and (5) \textbf{Few-Shot Baseline} (\cref{tab:fs_prompt_template}), which uses only example prompts without explicit persona context.(6) (\cref{tab:persona_rewrite_template}) demonstrates a fully instantiated persona with demographic, interest, and stylistic attributes.

\begin{table*}[ht]
    \noindent\fbox{\begin{minipage}{0.98\textwidth}
    \vspace{2mm}
    \textbf{Prompt Template for Persona Extraction from Posts and Comments}
    \vspace{2mm}
    \hrule
    \vspace{3mm}

    \colorbox{blue!5}{\parbox{0.96\columnwidth}{
    \textbf{\textcolor{blue}{System Prompt:}}

    You are an expert social media analyst who interprets patterns in people's posts and comments to form a realistic picture of who they might be.
    }}

    \vspace{3mm}

    \colorbox{orange!5}{\parbox{0.96\columnwidth}{
    \textbf{\textcolor{orange}{User Prompt:}}

    Below is a collection of a user's posts and comments:

    \quad\texttt{\{input\_data\}}

    \vspace{2mm}
    \textbf{Task}

    Read the posts and comments and infer what kind of person this user might be. Focus only on patterns that appear consistently across the data and ignore all NSFW content.

    Write a short persona description that captures the person's likely background, interests, and how they tend to express themselves online.

    \vspace{2mm}
    \textbf{What to infer when possible}

    Consider signals that may suggest:
    \begin{itemize}
    \item approximate age range
    \item gender (only if clearly suggested)
    \item location or region (only if implied)
    \item profession, field of study, or life context
    \item recurring interests, hobbies, or creative pursuits
    \item values, worldview, or themes that appear repeatedly
    \item communication style, tone, and typical ways of interacting
    \end{itemize}

    Avoid strong assumptions when evidence is weak.

    \vspace{2mm}
    \textbf{Output Format}

    Write a single paragraph in second person that begins with:

    \quad ``You are a [age range] [gender if inferable] from [location if implied] who...''

    The paragraph should read like a natural description of a person. Mention their background, interests, everyday environment, and the kinds of topics they discuss online. It may include small contextual details when they appear consistently (for example pets, places, routines, or cultural references).

    Do not include bullet points, lists, explanations of reasoning, references to posts or comments, or mentions of platforms or communities.

    The tone should be observational and grounded, as if describing a real person.
    }}
    
    \vspace{2mm}
    \end{minipage}}
    \caption{Prompt template for generating a persona from a user's posts and comments.}
    \label{tab:persona_extraction_template}
    \end{table*}

\begin{table*}[ht]
\noindent\fbox{\begin{minipage}{0.98\textwidth}
\vspace{2mm}
\textbf{Prompt for Zero-Shot Prompt Rewriting (Without Persona)}
\vspace{2mm}
\hrule
\vspace{3mm}

\textbf{User Prompt:}

You want to generate an image using a text to image model. A strong prompt clearly specifies the scene, subjects, visual details, mood, style, and lighting. Given a basic prompt, rewrite it into a richer and more vivid version that improves the resulting image.

\vspace{2mm}
The basic prompt is: \texttt{\{prompt\}}

\vspace{2mm}
Output only the enhanced prompt in fewer than 70 words. Do not include explanations or additional commentary.

\vspace{2mm}
\end{minipage}}
\caption{Prompt template for zero shot prompt rewriting without persona.}
\label{tab:zs_prompt_template}
\end{table*}

\begin{table*}[ht]
\noindent\fbox{\begin{minipage}{0.98\textwidth}
\vspace{2mm}
\textbf{ Prompt for Zero Shot Personalized Prompt Rewriting}
\vspace{2mm}
\hrule
\vspace{3mm}

\colorbox{blue!5}{\parbox{0.96\textwidth}{
\textbf{System Prompt:}

\texttt{\{Roleplay persona\}}
}}

\vspace{3mm}

\colorbox{orange!5}{\parbox{0.96\textwidth}{
\textbf{User Prompt:}

You want to generate an image using a text to image model. A strong prompt clearly specifies the scene, objects, visual details, mood, and style. You should rewrite the basic prompt so that it reflects your preferences, style, and personality. Your rewritten prompt must preserve the primary subjects and intent of the original prompt while aligning with your traits, personality and interests.

\vspace{2mm}
\textit{Basic Prompt:} \texttt{\{prompt\}}

\vspace{2mm}
Output only the rewritten prompt. Do not include explanations or commentary.
}}

\vspace{2mm}
\end{minipage}}
\caption{Prompt template for zero shot personalized prompt rewriting.}
\label{tab:zs_personalized_prompt_template}
\end{table*}

\clearpage

\begin{table*}[ht]
\noindent\fbox{\begin{minipage}{0.98\textwidth}
\vspace{2mm}
\textbf{Prompt for Few Shot Personalized Prompt Rewriting}
\vspace{2mm}
\hrule
\vspace{3mm}

\colorbox{blue!5}{\parbox{0.96\textwidth}{
\textbf{System Prompt:}

\texttt{\{Roleplay persona\}}
}}

\vspace{3mm}

\colorbox{orange!5}{\parbox{0.96\textwidth}{
\textbf{User Prompt:}

You want to generate an image using text to image generation. A strong prompt clearly specifies the scene, objects, visual details, mood, and style. Below are examples of prompts you have previously written, which reflect your preferences, artistic style, and characteristic choices:

\quad\textit{\{previous prompts\}}

\vspace{2mm}
Using these examples and your persona traits (interests, personality, and style), rewrite the given basic prompt so that it is consistent with both your persona and your demonstrated preferences. The rewritten prompt must retain the primary objects in the original prompt while expressing your distinctive stylistic choices.

\vspace{2mm}
\textit{Basic Prompt:} \texttt{\{prompt\}}

\vspace{2mm}
Output only the rewritten prompt. Do not include explanations or commentary.
}}

\vspace{2mm}
\end{minipage}}
\caption{Prompt template for few shot personalized prompt rewriting.}
\label{tab:fs_personalized_prompt_template}
\end{table*}

\clearpage

\begin{table*}[ht]
\noindent\fbox{\begin{minipage}{0.98\textwidth}
\vspace{2mm}
\textbf{Prompt for Few Shot Prompt Rewriting (Without Persona)}
\vspace{2mm}
\hrule
\vspace{3mm}

\textbf{User Prompt:}

You want to generate an image using text to image generation. A strong prompt clearly specifies the scene, objects, visual details, mood, and style. Below are examples of prompts the user has previously written, which reflect their preferences, artistic style, and characteristic choices:

\quad\textit{\{previous prompts\}}

\vspace{2mm}
Using these examples, rewrite the given basic prompt so that it is consistent with your demonstrated preferences. The rewritten prompt should retain the primary objects from the original prompt while adjusting the description to reflect the user's preferences as seen from their previous prompts.

\vspace{2mm}
\textit{Basic Prompt:} \texttt{\{prompt\}}

\vspace{2mm}
Output only the rewritten prompt. Do not include explanations or commentary.

\vspace{2mm}
\end{minipage}}
\caption{Prompt template for few shot prompt rewriting without persona.}
\label{tab:fs_prompt_template}
\end{table*}

\begin{table}[ht]
\noindent\fbox{\begin{minipage}{0.98\columnwidth}
\vspace{2mm}
\textbf{Prompt for Few Shot Personalized Prompt Rewriting (Detailed Example)}
\vspace{2mm}
\hrule
\vspace{3mm}

\colorbox{blue!5}{\parbox{0.96\columnwidth}{
\textbf{\textcolor{blue}{System Prompt:}}
You are a woman in your 30s–40s living in Los Angeles, originally from coastal Maine. You work in a hospital psychiatric setting and also explore art in your free time. You spend time online creating and discussing ideas about art, technology, science, and social issues. Your work explores imaginative and speculative themes, often touching on doomism, strange futures, and cultural commentary. Growing up along Maine’s foggy coast, pine forests, cliffs, and small towns shaped your sensibility. You are drawn to quiet, atmospheric environments and often reference landscapes, weather, and coastal life. You are curious about the strange and unexplained. Paranormal stories, folklore, and unusual phenomena interest you, though you often approach them with curiosity rather than certainty. You have a pet dog.
}}

\vspace{3mm}

\colorbox{orange!5}{\parbox{0.96\columnwidth}{
\textbf{\textcolor{orange}{User Prompt:}}

You want to generate an image using text to image generation. A strong prompt clearly specifies the scene, objects, visual details, mood, and style. Below are examples of prompts you have previously written, which reflect your preferences, artistic style, and characteristic choices:

\quad\textit{\{previous prompts\}}

\vspace{2mm}
Using these examples and your persona traits (interests, personality, and style), rewrite the given basic prompt so that it is consistent with both your persona and your demonstrated preferences. The rewritten prompt must retain the primary objects in the original prompt while expressing your distinctive stylistic choices.

\vspace{2mm}
\textit{Basic Prompt:} \texttt{\{prompt\}}

\vspace{2mm}
Output only the rewritten prompt. Do not include explanations or commentary.
}}

\vspace{2mm}
\end{minipage}}
\caption{Persona-integrated prompt template for few-shot personalized prompt rewriting with fully instantiated persona attributes}
\label{tab:persona_rewrite_template}
\end{table}

\clearpage

\section{Dataset Details}
We introduce two datasets in this paper, the Reddit Interactions bipartite graph which after thorough filtering (explained below in \Cref{sec:supp-filtering}) consists of 23M users, 682M posts and comments, and 382k images posted by them across 40k subreddits. On creating a user intersection across platforms on Civitai, we construct \zipbench which consists of 1.5k users, with 198k posts and comments and 40k images generated by them on Civitai.

\begin{table}[ht]
\centering
\begin{tabular}{lccc}
\toprule
Dataset & Users & Images & Posts and Comments \\
\midrule
ZIP Bench & 1.5K & 40K & 198K \\
Reddit Interactions & 23M & 382K & 682M \\
\bottomrule
\end{tabular}
\caption{Statistics for the two datasets used in the paper.}
\end{table}
\label{sec:supp-data}
\vspace{-5em}
\subsection{Reddit interactions} 
\paragraph{Filtering and deduplication.}
\label{sec:supp-filtering} For cleaning and curating a high quality dataset we adapt the dataset filtering method proposed by BehaviorLLaVA ~\cite{singh2025teaching}. We use the same methodologies for the following steps to filter out noisy data.
\begin{enumerate}
    \item exclude bot-like accounts and automated moderation accounts
    \item filter NSFW and toxic content using LlaMA Guard
    \item drop posts/comments from ``[deleted]'' users
    \item We filter out comments that are less than 5 words
\end{enumerate}

Further we strip private identifiers such as URLs, names, and exact geolocation. We also remove explicit sensitive attributes including political or religious declarations, sexual orientation, and medical information, to prevent such attributes from being encoded in the persona. To avoid skew from power users or dominant communities, we cap per-user and per-subreddit contribution volumes and downweight extremely active accounts or highly popular subreddits that would otherwise overshadow finer-grained interests. Finally, we remove ``low-information’’ users whose histories contain very sparse, repetitive, or semantically empty content, as these interactions provide insufficient behavioral signal.

Subreddits and Users form the classes of nodes, and edges indicate a post or comment, the edge weight being the 2 length vector [number of comments, number of posts]. Leaving us with the bipartite graph described above.

\vspace{-1em}
\subsection{Linking Civitai Users to Reddit Personas}
\label{sec:supp-civitai-linking}
When users login Civitai using their reddit account say \textbf{\textit{``u/ACivitaiGuy"}}\footnote{The account username is fictional and used only for illustration}, their usernames are the same as their \url{https://civitai.com/user/ACivitaiGuy} or \url{https://civitai.com/user/ACivitaiGuy_XXX} where XXX is a 3 digit number (if the username already exists). Using the Civitai user-profile API, we enumerate roughly 100K users and identify 13K with at least five safe for work shared images. Applying the url matching and alias logic above, we identify 1.5k users, with 40k Civitai generations from the Civitai API. From reddit we have 31K posts and 600K comments for these users.
 
\subsection{Dataset Distribution}
\begin{figure}[H]
    \centering
    
    \begin{subfigure}[t]{0.49\linewidth}
        \centering
        \includegraphics[width=\linewidth]{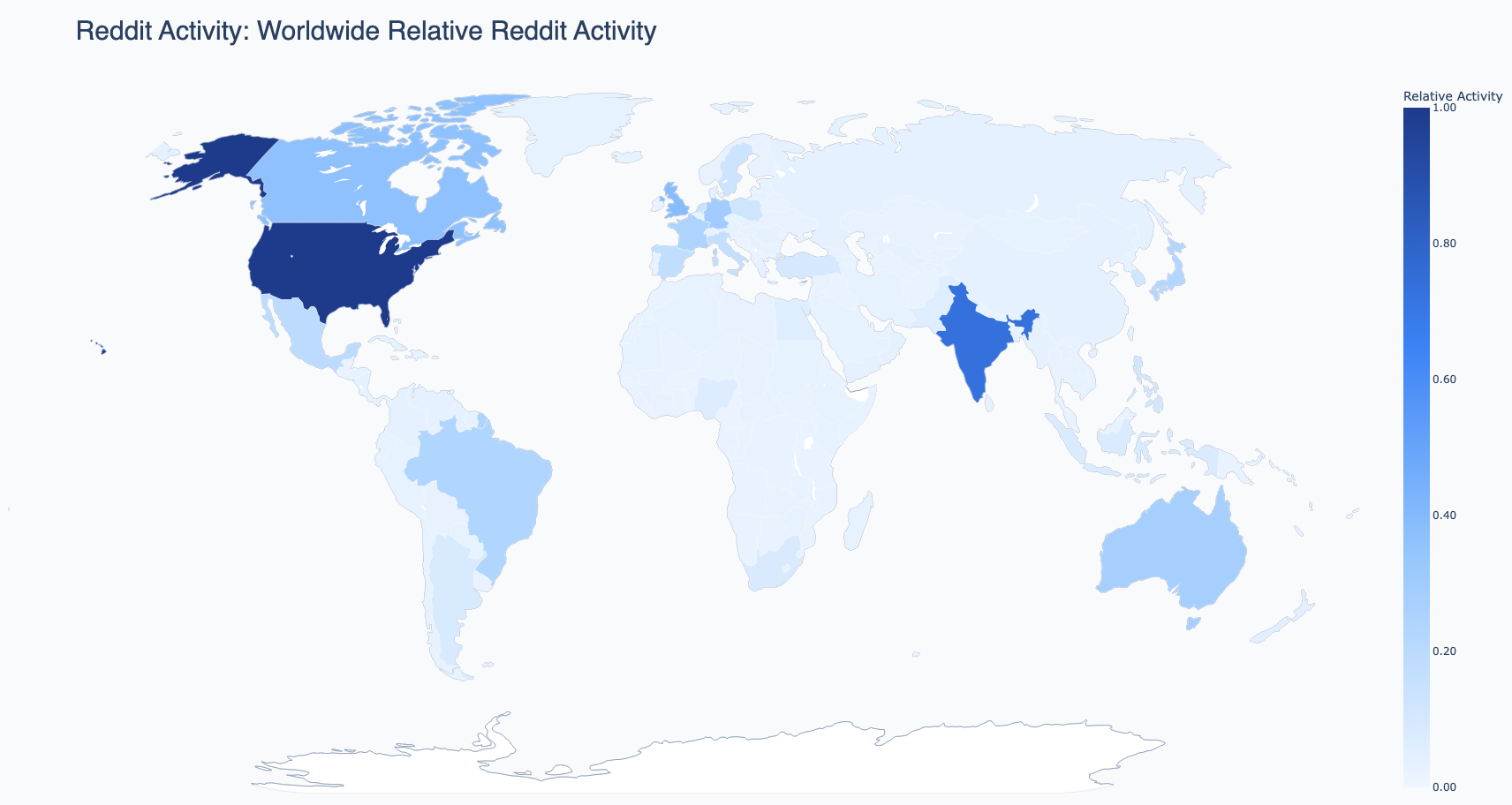}
        \caption{Heatmap of users in our \zipbench across the world}
        \label{fig:world_heatmap}
    \end{subfigure}
    \hfill
    \begin{subfigure}[t]{0.49\linewidth}
        \centering
        \includegraphics[width=\linewidth]{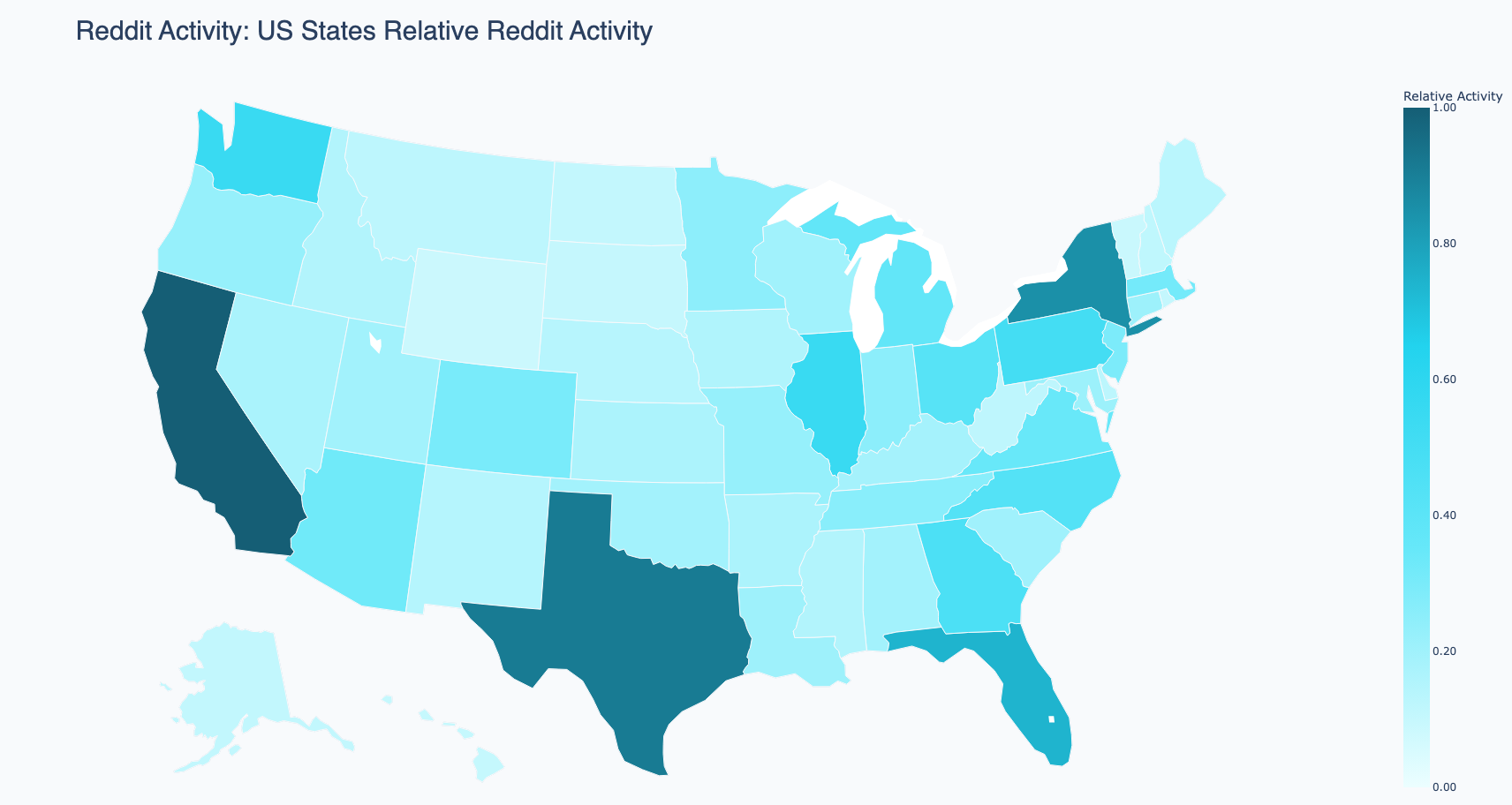}
        \caption{Heatmap of users in \zipbench dataset across USA}
        \label{fig:usa_heatmap}
    \end{subfigure}

    \vspace{0.6em}

    \begin{subfigure}[t]{0.8\linewidth}
        \centering
        \includegraphics[width=\linewidth]{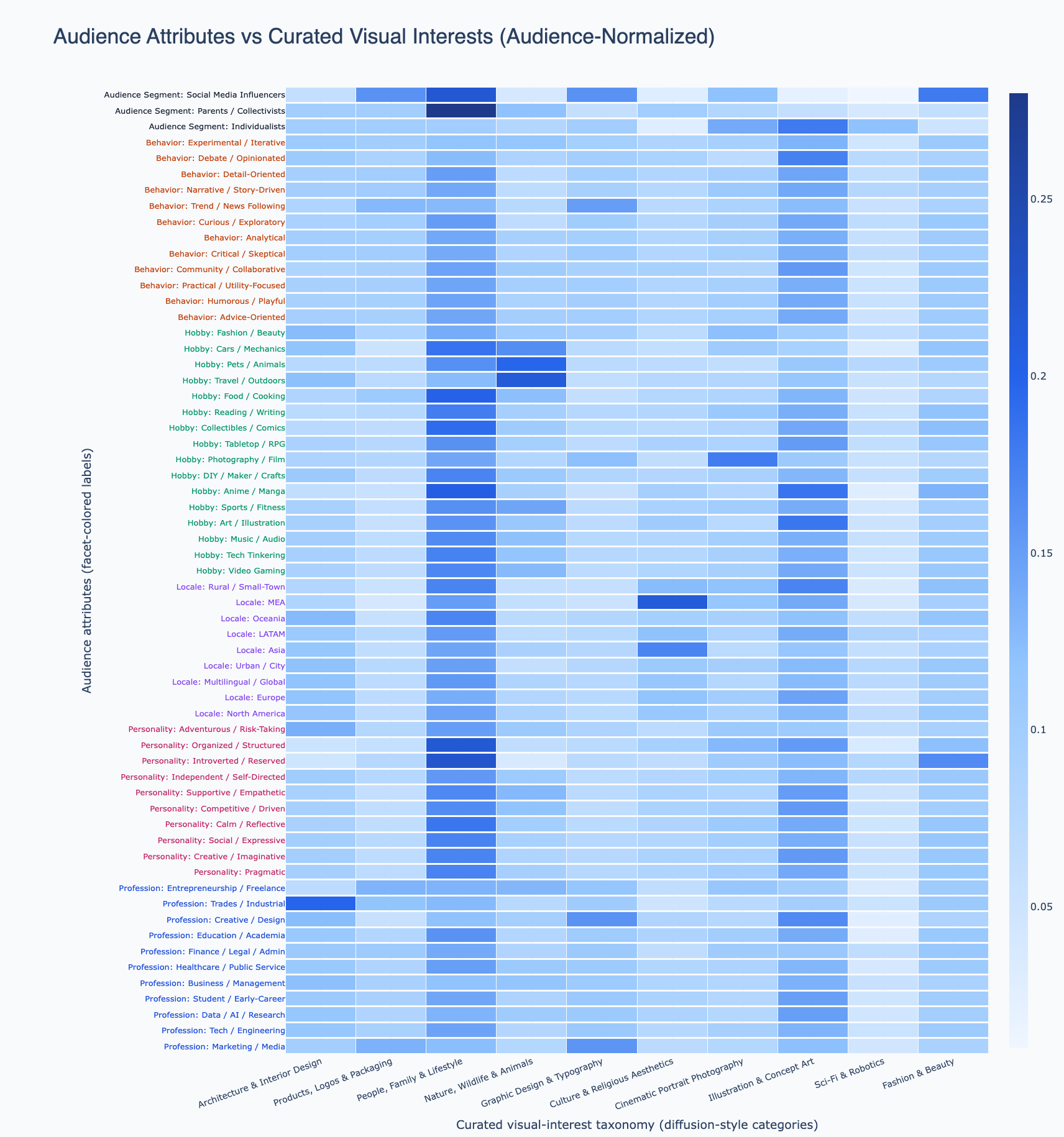}
        \caption{Heatmap of visual interests (professions, behavior, locale, personality, hobby) across persona segments in \zipbench}
        \label{fig:persona_heatmap}
    \end{subfigure}

    \caption{Geographic and persona-level heatmaps for the \zipbench dataset.}
    \label{fig:zipbench_heatmaps}
\end{figure}
\subsubsection{Demographics}
\label{sec:supp-demographics}
We profile the demographic composition of \zipbench using personas mined from Reddit activity. Since ground-truth demographics are unavailable, we rely on GPT-4o--inferred attributes extracted during persona verbalization (\cref{sec:supp-persona}). We report distributions across region and profession.
\paragraph{Region.}
North America dominates (38\%), followed by Western Europe (24\%), South Asia (12\%), and East Asia (11\%). Latin America (7\%), Africa (5\%), and the Middle East/CIS region (3\%) are underrepresented. Geographic heatmaps at the country and U.S.\ state level are shown in \cref{fig:world_heatmap,fig:usa_heatmap}.

\paragraph{Profession.}
STEM and engineering professionals constitute the largest group (30\%), followed by creative/arts (18\%), students (15\%), and business/finance (14\%). Education (10\%), healthcare (8\%), and manual trades (5\%) round out the distribution. A cross-tabulation of persona attributes (profession, hobbies, locale, personality) is visualized in \cref{fig:persona_heatmap}.

These skews motivate our use of IPF reweighting (\cref{alg:ipf_reweighting}) to produce demographically balanced aggregate scores, and our per-subgroup reporting in \cref{tab:subgroup_cs} to surface performance disparities that raw aggregates would mask.

\subsection{Degree distributions}
\label{sec:supp-degree-dist}
It is typical for social networks to exhibit Poisson Distribution on node degrees, the problem with representation learning on such graphs is aggregation to central nodes (high-activity subreddits / power users), therefore we visualize the log-log relationship and see that log normalized distributions are close to Gaussians, therefore we use the Power Law normalization similar to PINSage.
\label{ssec:reddit-analysis}
\label{sec:supp-degree}
\begin{figure}[ht]
    \centering
    \begin{subfigure}[t]{0.48\linewidth}
        \centering
        \includegraphics[width=\linewidth]{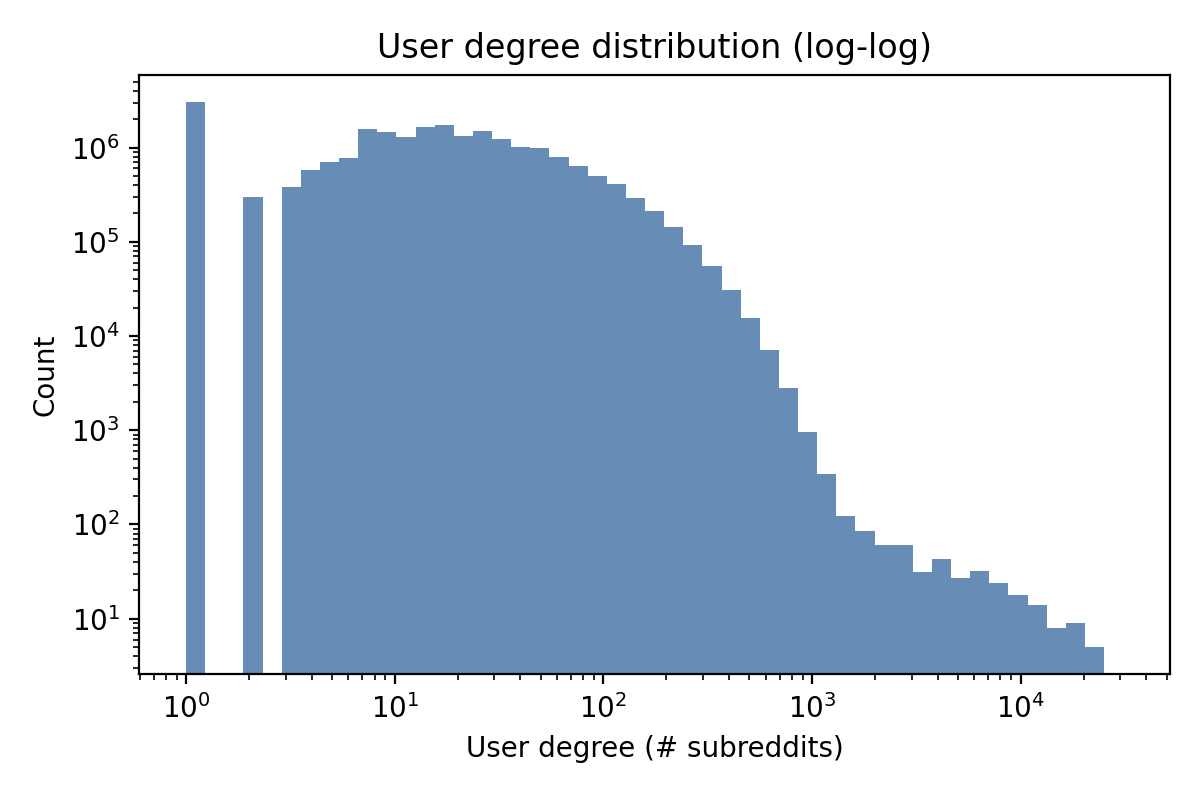}
        \caption*{(a) User degree distribution (log--log).}
        \label{fig:deg-users}
    \end{subfigure}
    \hfill
    \begin{subfigure}[t]{0.48\linewidth}
        \centering
        \includegraphics[width=\linewidth]{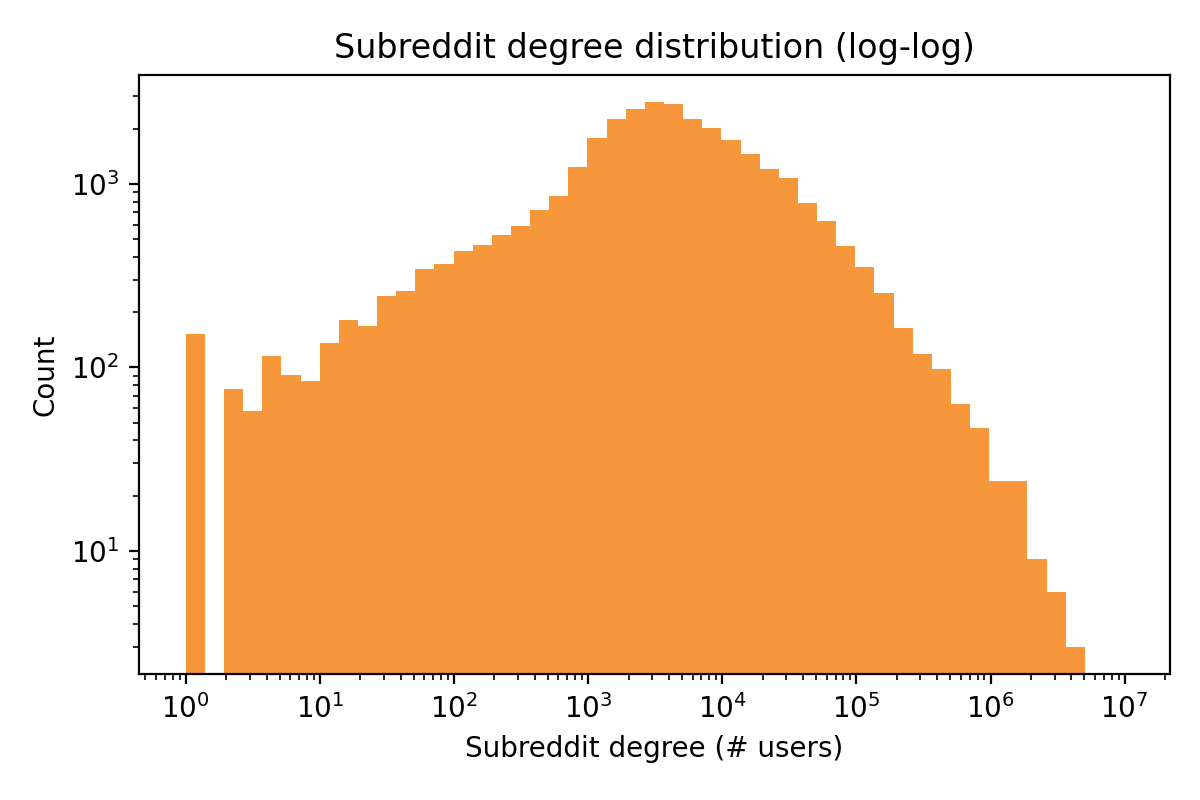}
        \caption*{(b) Subreddit degree distribution (log--log).}
        \label{fig:deg-subreddits}
    \end{subfigure}
    \caption{Reddit Interactions Graph Statistics show a poisson distribution (normal under log log), alluding to our choices for edge weight normalization}
\end{figure}

\section{Experiments \& Results}
\label{sec:supp-results}

\subsection{Graph ablations}
\label{sec:supp-gnn-ablations}

A natural question is whether graph-mined personas truly reflect user identity or merely correlate with surface-level activity. We address this through three complementary ablation studies: (i)~encoder architecture and supervision objectives, (ii)~popularity bias mitigation via subreddit filtering, and (iii)~architectural hyperparameters.

\subsubsection{Encoder architecture and supervision}
We evaluate all methods for persona verbalization on posting accuracy and our ZIP Lift metric. As shown in \cref{tab:graph_ablation}, \textbf{GAT + ImageAlign} reaches \textbf{73\%} posting accuracy, demonstrating strong user preference and behavior modeling capabilities.

The personalization metrics support the same conclusion. ZIP Lift scales with behavioral fidelity: \textbf{GAT + ImageAlign} yields a \textbf{17.7\%} improvement, compared to only \textbf{5--6\%} for TF-IDF baselines. This demonstrates that richer and more behaviorally aligned personas directly translate into stronger zero-shot personalization.

Two design choices are essential.
\textbf{Attention instead of TF-IDF}: Learned edge-aware attention in GAT and GraphSAGE consistently outperforms TF-IDF heuristics, raising ZIP Lift to the 8--12\% range. These gains indicate better identification of behaviorally salient subreddits.
\textbf{ImageAlign supervision}: Adding the ImageAlign task improves every architecture by 3--11\% in ZIP Lift. This confirms that grounding user embeddings in visual posting behavior captures aspects of persona that are not available from graph structure alone.

\subsubsection{Popularity bias and subreddit filtering}
\label{sec:supp-subreddit-filtering}
Social interaction graphs exhibit heavy-tailed degree distributions (\cref{sec:supp-degree}): a small number of mega-subreddits (e.g., r/AskReddit, r/funny) account for a disproportionate share of total activity, while the bottom 12k subreddits collectively represent only 0.1\% of all interactions. Without mitigation, the GATv2 attention mechanism converges toward these high-degree nodes, producing user embeddings that reflect generic popularity rather than distinctive preferences.

We apply two complementary strategies.
\textbf{(1) Degree-normalized edge weights.} As described in \cref{sec:supp-gnn}, we normalize raw interaction counts via $w_{us} = \log(1 + t_{us}) / \sqrt{d_u \, d_s + \varepsilon}$, which penalizes high-degree nodes on both the user and subreddit sides. This prevents power users and dominant communities from overwhelming finer-grained interests during message passing.
\textbf{(2) Long-tail subreddit pruning.} We remove the bottom-$K$ subreddits ranked by total user interactions before graph construction. The pruned communities are predominantly unmoderated, frequently banned, or lack sufficient content for meaningful behavioral signal; retaining them introduces noise into the learned attention weights.

To determine the optimal filtering threshold, we sweep over different values of $K$ and measure the impact on both graph-level training metrics (posting accuracy) and downstream personalization (ClipScore on \zipbench). \Cref{tab:subreddit_ablation} reports the results. On \zipbench, only \textbf{1 user} appears in the filtered subreddits, indicating negligible direct overlap; the effects instead propagate through changes in learned attention distributions.

\begin{table}[t]
\centering
\small
\renewcommand{\arraystretch}{1.15}
\setlength{\tabcolsep}{8pt}
\begin{tabular}{@{}lcc@{}}
\toprule
\textbf{Subreddits retained} & \textbf{$\Delta$ Posting Acc.\ (\%)} & \textbf{$\Delta$ ClipScore (\%)} \\
\midrule
5k   & $-$11.7 & $-$2.1 \\
10k  & $-$5.0  & $-$2.0 \\
15k  & $+$0.1  & $+$0.3 \\
20k  & $+$0.4  & $-$4.5 \\
\bottomrule
\end{tabular}
\caption{\textbf{Effect of subreddit filtering threshold.} Deltas are relative to the full graph (all $\sim$32k subreddits). Retaining 12k--15k subreddits yields the best trade-off: aggressive pruning (5k) removes behaviorally useful niche communities, while retaining too many (20k) introduces noisy, long-tailed subreddits that degrade downstream personalization despite marginal gains in posting accuracy.}
\label{tab:subreddit_ablation}
\end{table}

Removing fewer than 12k subreddits retains noisy long-tail communities that degrade ClipScore ($-$4.5\% at 20k), while removing more than 15k discards niche but behaviorally rich communities, hurting posting accuracy ($-$11.7\% at 5k). The 12k--15k regime balances graph training quality with downstream personalization, and is used for all experiments in the paper.

\subsubsection{Architectural hyperparameters}
\label{sec:supp-arch-ablation}
\begin{figure*}[t]
\centering

\begin{subfigure}[t]{0.32\linewidth}
    \centering
    \includegraphics[width=\linewidth]{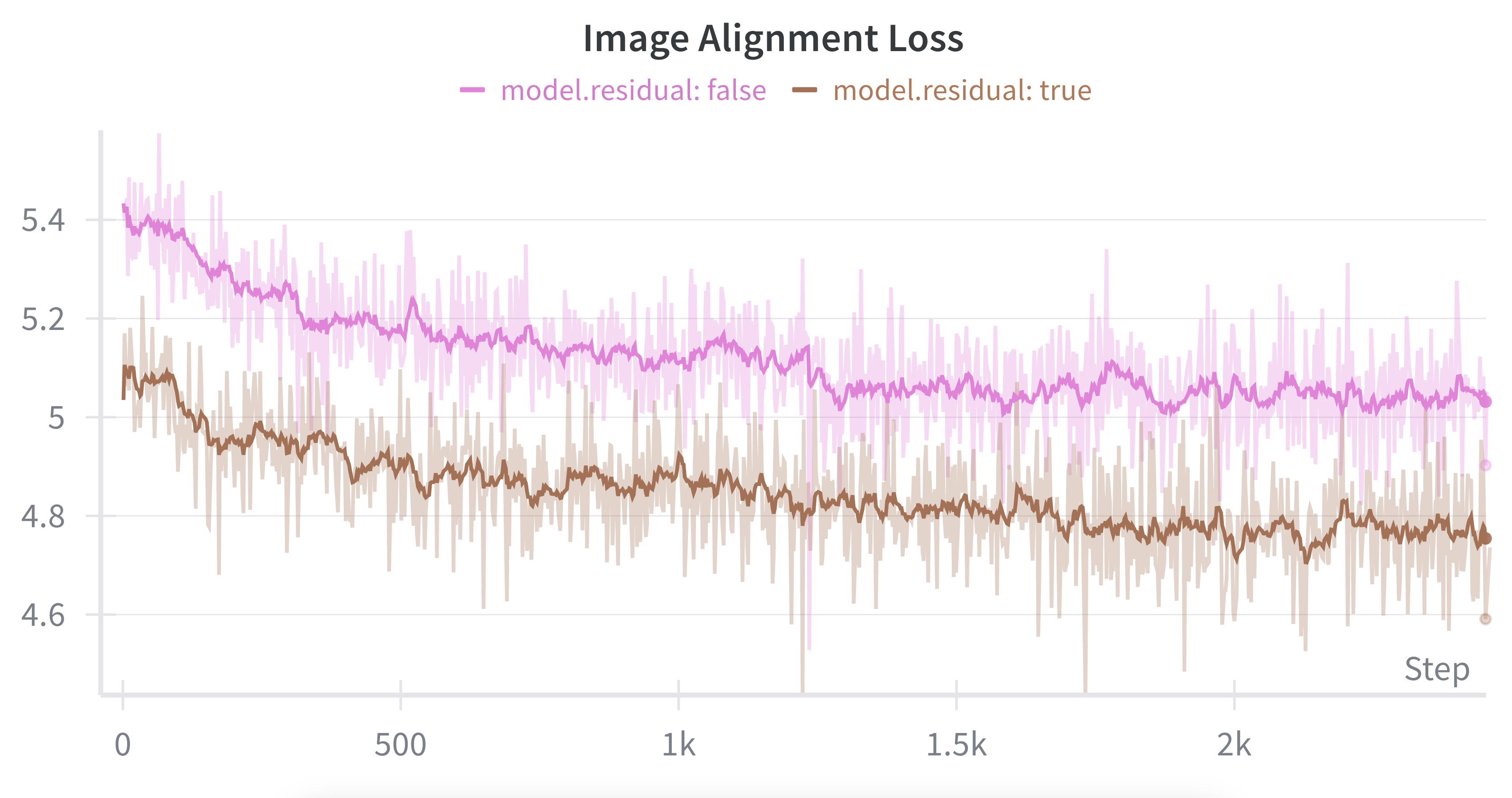}
    \caption{\textbf{Residuals Improve Image Alignment.}
    Enabling residual links between GATv2 layers improves convergence stability and lowers alignment loss compared to the non-residual variant.}
    \label{fig:residual_image_alignment}
\end{subfigure}
\hfill
\begin{subfigure}[t]{0.32\linewidth}
    \centering
    \includegraphics[width=\linewidth]{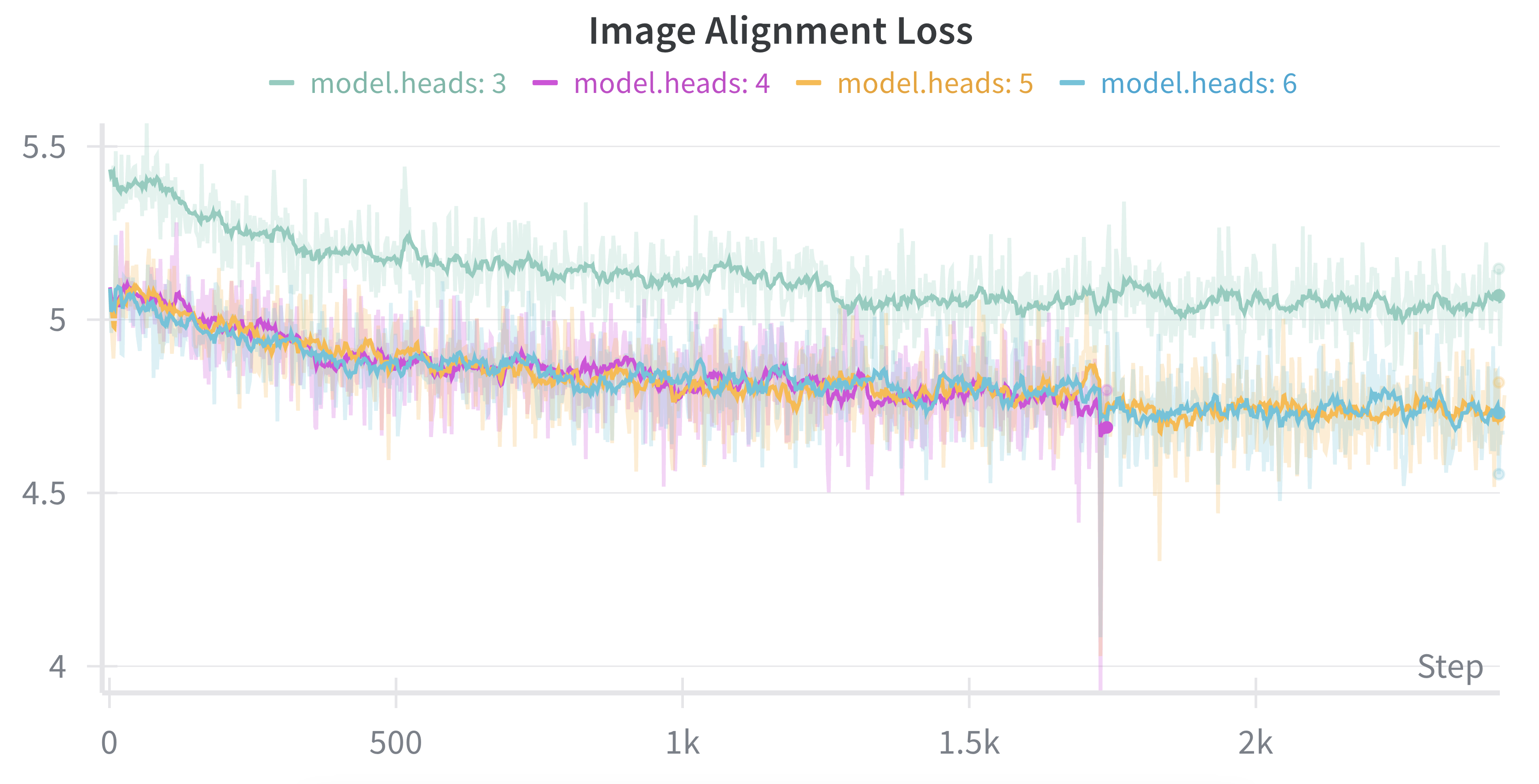}
    \caption{\textbf{Attention Head Ablation.}
    Increasing the number of attention heads improves expressivity; four to six heads converge faster and reach lower loss.}
    \label{fig:heads_image_alignment}
\end{subfigure}
\hfill
\begin{subfigure}[t]{0.32\linewidth}
    \centering
    \includegraphics[width=\linewidth]{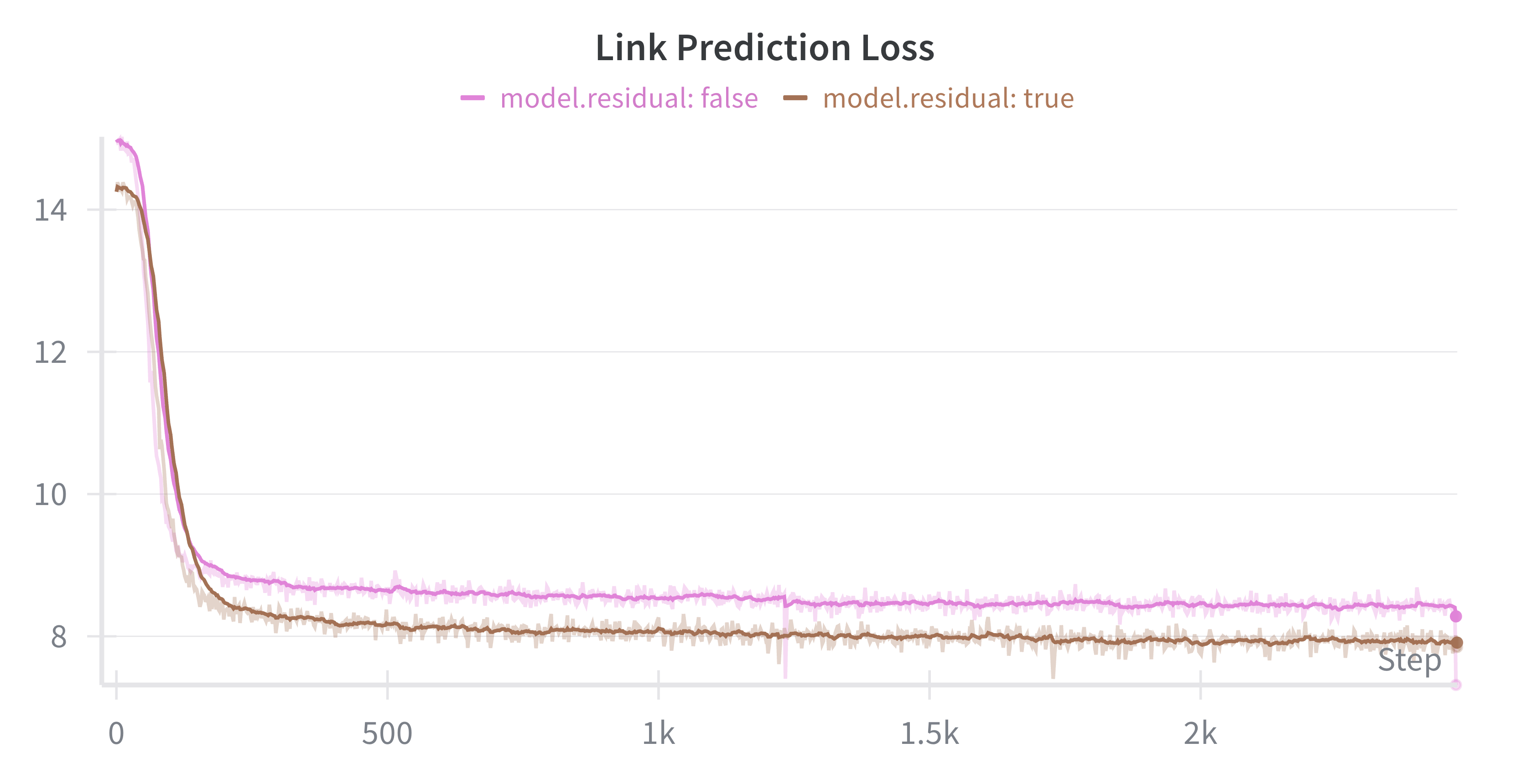}
    \caption{\textbf{Residuals Improve Link Prediction.}
    Skip connections accelerate convergence and stabilize link prediction loss.}
    \label{fig:residual_link_prediction}
\end{subfigure}

\caption{
\textbf{Training Dynamics of the Edge-Aware GATv2 Encoder.}
Residual connections significantly stabilize optimization and improve both image–embedding alignment and structural link prediction. Additionally, moderate multi-head attention improves representational capacity without sacrificing optimization stability.
}

\label{fig:training_dynamics_ablation}

\end{figure*}

In \cref{fig:heads_image_alignment,fig:residual_image_alignment,fig:residual_link_prediction} we report training curves for different architectural choices.

\textbf{Residual connections.}
Enabling residual links between GATv2 layers substantially improves convergence stability and lowers both the image alignment loss (\cref{fig:residual_image_alignment}) and the link prediction loss (\cref{fig:residual_link_prediction}). Residual propagation mitigates over-smoothing and preserves high-frequency user--subreddit signals across message-passing steps.

\textbf{Number of attention heads.}
Increasing from 3 to 4 heads yields a clear improvement in alignment loss, but adding further heads (5--6) provides no additional gain (\cref{fig:heads_image_alignment}). We therefore use $K = 4$ heads for all experiments, balancing representational diversity with optimization stability.

\begin{table}[hbtp!]
\centering
\small
\renewcommand{\arraystretch}{1.15} %
\setlength{\tabcolsep}{6pt}

\begin{adjustbox}{max width=\columnwidth}
\begin{tabular}{@{}l cc@{}}
\toprule
\textbf{Method} & \textbf{Posting Acc.} $\uparrow$ & \textbf{ZIP Lift (\%)} $\uparrow$ \\
\midrule
\multicolumn{3}{l}{\textit{Statistical Baselines (w/ GPT-4o)}} \\
TF-IDF & 55 & 5.1 \\
TF-IDF + NMI & 57 & 5.3 \\

\midrule
\multicolumn{3}{l}{\textit{Graph Encoders (w/ GPT-4o)}} \\
LightGCN & 59 & 6.2 \\
\hspace{1em} $\hookrightarrow$ + ImageAlign & 62 & 9.1 \\
\addlinespace
GraphSAGE & 63 & 8.5 \\
\hspace{1em} $\hookrightarrow$ + ImageAlign & 67 & 10.2 \\
\addlinespace
GAT & 62 & \valgood{11.8} \\
\hspace{1em} $\hookrightarrow$ + ImageAlign & \valbest{73} & \valbest{17.7} \\

\bottomrule
\end{tabular}
\end{adjustbox}

\caption{\textbf{Comparison of personalization methods.}
We evaluate statistical and graph-based approaches (paired with GPT-4o).
Rows marked with $\hookrightarrow$ indicate the addition of the auxiliary \textit{ImageAlign} task.
\valbest{Bold} indicates best performance; \valgood{Underline} indicates second best.
Adding ImageAlign consistently improves all metrics, with GAT + ImageAlign achieving the highest ZIP Lift.}
\label{tab:graph_ablation}
\end{table}

\begin{table}[t]
\centering
\setlength{\tabcolsep}{8pt}
\renewcommand{\arraystretch}{1.2}
\begin{tabular}{l l c}
\toprule
\textbf{Model} & \textbf{Method} & \textbf{CLIPScore $\uparrow$} \\
\midrule

\multirow{4}{*}{SD XL} 
& -       & 58.15 \\
& FABRIC  & 64.15 \\
& TV      & 63.60 \\
& DrUM & 66.93 \\
& \textbf{ZIPPY(5-shot)} & \textbf{68.65} \\
\midrule

\multirow{4}{*}{SD V3}
& -       & 56.50 \\
& FABRIC  & 59.22 \\
& TV      & 59.63 \\
& DrUM & 62.85 \\
& \textbf{ZIPPY(5-shot)} & \textbf{64.39} \\
\midrule

\multirow{4}{*}{SD V3.5} 
& -       & 57.00 \\
& FABRIC  & 61.33 \\
& TV      & 62.45 \\
& DrUM & 64.18 \\
& \textbf{ZIPPY(5-shot)} & \textbf{67.12} \\
\bottomrule
\end{tabular}
\caption{Cross-model generalization of personalization methods. Our persona-based approach consistently outperforms baselines across different Stable Diffusion versions, with the largest gains on SD-XL (up to $+4.5\%$ CLIPScore improvement), demonstrating robustness to underlying generation model changes. CLIP score is calculated between the personalized image and the original prompt by the user.}
\end{table}

\FloatBarrier

\subsection{Zero shot results}
\label{sec:supp-zero-shot}

\begin{table}[p]
\centering
\begin{adjustbox}{width=0.65\linewidth}
\begin{tabular}{lcc}
\toprule
\textbf{Model} & \multicolumn{2}{c}{\textbf{CLIPScore $\uparrow$}} \\
 & \textbf{w/o P} & \textbf{w/ P (\%$\uparrow$)} \\
\midrule
Qwen3-8B & 0.60 & 0.62 (+3\%) \\
\textbf{Qwen3-32B} & 0.61 & 0.69 (+13\%) \\
Qwen3-30B-A3B & 0.58 & 0.61 (+5\%) \\
Qwen3-235B-A22B & 0.61 & 0.67 (+10\%) \\
\midrule
GPT-4o & 0.62 & 0.69 (+11\%) \\
GPT-4.1 & 0.62 & 0.67 (+8\%) \\
\textbf{GPT-5-chat} & 0.62 & \valgood{0.72 (+16\%)} \\
\midrule
Claude-3-Opus & 0.62 & 0.72 (+16\%) \\
Claude-Sonnet-3.7 & 0.62 & 0.70 (+13\%) \\
\textbf{Claude-Sonnet-4} & 0.63 & \valbest{0.76 (+20\%)} \\
\midrule
DeepSeek-V3 & 0.62 & 0.66 (+6\%) \\
\textbf{DeepSeek-V3.2-Exp} & 0.62 & 0.67 (+8\%) \\
\midrule
Gemini 2.5 Flash & 0.61 & 0.66 (+8\%) \\
\textbf{Gemini 2.5 Pro} & 0.62 & 0.68 (+10\%) \\
\bottomrule
\end{tabular}
\end{adjustbox}
\caption{\textbf{Zero-shot personalization scores across LLMs.}
Persona conditioning (w/ P) yields consistent improvements across all model families, with relative gains from +3\% to +20\%.
Claude-Sonnet-4 achieves the highest score (0.76 CLIP), followed by GPT-5-chat (0.72) and Qwen3-32B (0.69).
The results show that persona conditioning serves as a universal and efficient zero-shot personalization signal, enhancing multimodal alignment without any fine-tuning.}
\label{tab:zip_clip_imagealign_table}
\end{table}

\vspace{-0.5em}\paragraph{Performance across 12 LLMs.} Persona conditioning yields consistent improvements across all evaluated models (\cref{tab:zip_clip_imagealign_table}), with relative gains from +3\% (Qwen3-8B, Qwen3-30B-A3B) to +20\% (Claude-Sonnet-4). Frontier models (Claude, GPT-5) achieve 13–20\% gains, substantially outperforming open-source alternatives (3–13\% for Qwen, 6–8\% for DeepSeek). We attribute this gap to stronger instruction-following capabilities validated in recent roleplay benchmarks~\cite{du2025twinvoice}, where Claude ranks highest. Mixture-of-experts models (Qwen3-30B-A3B, Qwen3-235B-A22B) underperform dense counterparts despite larger parameter counts—manual inspection reveals repetitive over-personalization, suggesting routing instability when conditioning on long persona contexts.

\FloatBarrier

\begin{figure*}[!t]
    \centering
    \includegraphics[width=1\linewidth]{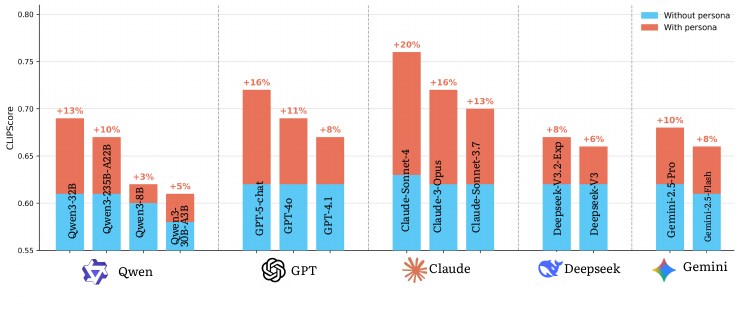}
    \caption{ClipScores by model family with persona conditioning. Stacked bars show baseline (blue) and improvement from persona conditioning (red). All models benefit from persona conditioning ranging from 3\% (Qwen3-30B-A3B) to 20\% (Claude-Sonnet-4). Frontier models achieve the strongest gains (13--20\%), outperforming open-source alternatives (3--13\%).}
\label{fig:suppl:imagealign_by_family}
\end{figure*}

\vspace{-0.5em}\paragraph{Token budget and context scaling.} Performance increases steadily up to 2048 tokens, after which most models saturate (\cref{fig:token-budget}). Claude-Sonnet-4 maintains gains beyond 4096 tokens, reaching 0.76 ImageAlign at 8192 tokens. Full personas (subreddits + comment snippets) outperform subreddit-only cues by 10–15\% at equivalent budgets, validating that textual engagement signals provide richer preference information than community membership alone. GPT-4o begins hallucinating persona details (fabricating subreddit memberships, inventing preferences) beyond 8192 tokens, motivating our 4096-token budget cap.

\subsection{Few-Shot Results}
\label{sec:supp-few-shot}

We provide extended few-shot results complementing the main evaluation in \cref{tab:main_results}. \Cref{fig:shot-lineplot} plots personalization performance (CLIPScore) as a function of the number of in-context examples ($n=0$--$5$) across representative models.

\paragraph{Few-shot scaling behavior.}
\Cref{fig:shot-lineplot} reveals divergent few-shot learning dynamics across model families. Open-source models (e.g., Qwen3-32B) exhibit rapid saturation: performance jumps sharply at $n{=}1$--$2$ examples and plateaus by $n{=}3$, with persona conditioning providing additive but diminishing gains beyond this point.
Frontier models (GPT-5, Claude-Sonnet-4, Gemini~2.5~Pro) display a markedly different trajectory, they continue to improve steadily up to $n{=}5$ shots, suggesting stronger in-context learning capacity that can leverage additional examples more effectively.
Across all models, persona conditioning shifts the entire curve upward, meaning the gains from persona priors and few-shot examples are largely complementary rather than redundant. Notably, a persona-conditioned model at $n{=}0$ often matches or exceeds a non-persona model at $n{=}3$, underscoring the data efficiency of persona priors.

\textit{Persona conditioning is universally beneficial.} Every model improves on all three metrics when conditioned on persona context. The average relative gain is +4.5\% on CLIPScore, and +4.1\% on ImageAlign, confirming that persona priors provide complementary signal regardless of model architecture or scale.

\subsection{Persona Token Budget}
\label{sec:supp-token-budget}
The persona token budget $B$ controls how much of the mined user profile is presented to the LLM during prompt rewriting. Smaller budgets retain only the highest-attention subreddits and top-ranked posts, while larger budgets include progressively more niche communities and comment excerpts. This trade-off is central to persona-conditioned personalization: too little context yields generic rewrites indistinguishable from an unpersonalized baseline, while excessive context introduces noise and risks hallucination (\cref{sec:supp-limitations}).

\begin{figure}[!t]
    \centering
    \includegraphics[width=0.5\linewidth]{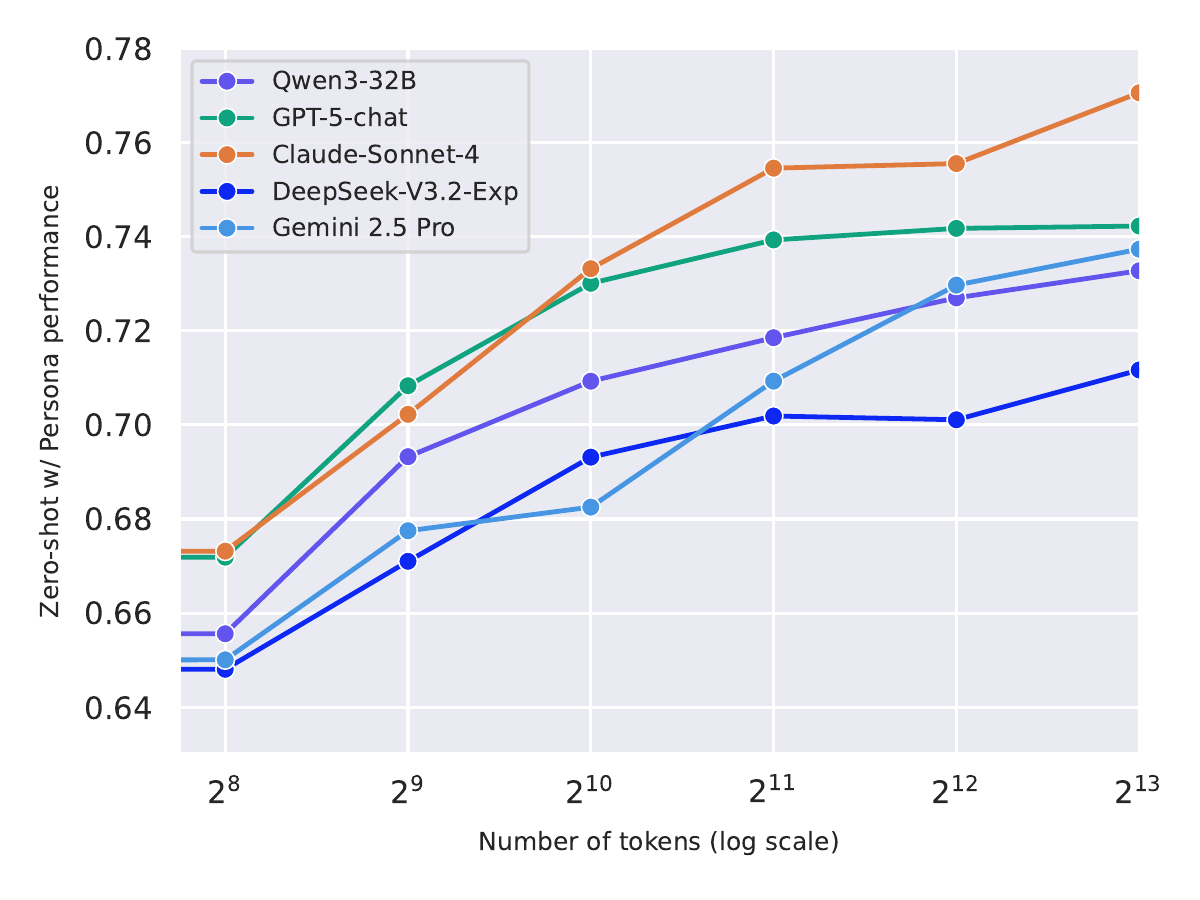}
    \caption{\textbf{Effect of persona token budget on ZIPP}. Zero-shot personalization CLIPScores (Y-axis) improve consistently with larger persona context budget B, showing diminishing returns beyond 2048 tokens.}
    \label{fig:token-budget}
\end{figure}

\Cref{fig:token-budget} shows CLIPScore as a function of $B$ for five representative models. All models follow a log-linear improvement up to $\sim$2048 tokens, after which returns diminish. We adopt $B{=}4096$ as the default across all experiments, as it consistently lies in the plateau region for the majority of models while remaining below the hallucination threshold observed for GPT-4o at 8192 tokens.

\subsection{Demographic alignment}
\label{sec:supp-generalization}
We present a per-subgroup breakdown of personalization performance on the RapidData benchmark as it provides explicitly recoreded demographics.
RapidData annotates each preference pair across five categorical axes: \textit{age}, \textit{gender}, \textit{country/region}, \textit{language}, and \textit{profession}.
We aggregate country and language into broader regional groupings for clarity and report per-subgroup CLIPScore in \Cref{tab:subgroup_cs}.
The ``Sample~\%'' column reports the fraction of RapidData users in each subgroup, exposing the skew that motivates IPF reweighting (\cref{ssec:pluralism}).

\FloatBarrier

\begin{table*}[!t]
\centering
\begin{adjustbox}{max width=\textwidth}
\setlength{\tabcolsep}{5pt}
\renewcommand{\arraystretch}{1.1}
\begin{tabular}{@{}lrcccccc@{}}
\toprule
\textbf{Subgroup} & \textbf{Samp.\,\%}
  & \textbf{LLM} & \textbf{TV\,(3)} & \textbf{DrUM} & \textbf{ViPer}
  & \textbf{\MyModel\,(0)} & \textbf{\MyModel\,(5)} \\
& & (0-shot) & (few-shot) & (FT) & (FT) & (0-shot) & (few-shot) \\
\midrule

\multicolumn{8}{l}{\textit{\textbf{Age}}} \\
\quad 18--24 & 28 & 68.2 & 78.0 & 72.0 & 74.3 & 74.2 & \textbf{78.5} \\
\quad 25--34 & 35 & 68.5 & \textbf{79.0} & 73.5 & 74.5 & 74.8 & \textbf{79.0} \\
\quad 35--44 & 18 & 67.5 & 77.5 & 70.5 & 74.0 & 73.8 & \textbf{78.5} \\
\quad 45--54 & 12 & 65.0 & 75.0 & 64.5 & 73.5 & 72.8 & \textbf{77.5} \\
\quad 55+ & 7 & 62.8 & 72.0 & 58.5 & 73.8 & 71.5 & \textbf{76.8} \\

\midrule
\multicolumn{8}{l}{\textit{\textbf{Gender}}} \\
\quad Male & 62 & 68.0 & 78.0 & 72.5 & 74.2 & 74.2 & \textbf{78.5} \\
\quad Female & 32 & 66.8 & 76.8 & 68.0 & 74.0 & 73.5 & \textbf{78.0} \\
\quad Non-binary / Other & 6 & 64.5 & 74.5 & 63.0 & 73.5 & 72.5 & \textbf{77.0} \\

\midrule
\multicolumn{8}{l}{\textit{\textbf{Region}}} \\
\quad N.\,America & 38 & 69.5 & \textbf{79.5} & 75.0 & 74.5 & 75.0 & 79.0 \\
\quad W.\,Europe & 24 & 68.5 & 78.5 & 73.0 & 74.2 & 74.5 & \textbf{78.8} \\
\quad S.\,Asia & 12 & 66.5 & 76.0 & 67.0 & 74.0 & 73.5 & \textbf{78.0} \\
\quad E.\,Asia & 11 & 67.0 & 76.5 & 68.0 & 74.0 & 73.8 & \textbf{78.2} \\
\quad Lat.\,America & 7 & 64.0 & 73.5 & 61.5 & 73.5 & 72.0 & \textbf{77.2} \\
\quad Africa & 5 & 62.5 & 71.8 & 58.0 & 73.5 & 71.0 & \textbf{76.5} \\
\quad ME\,/\,CIS & 3 & 63.5 & 72.5 & 60.0 & 73.5 & 71.5 & \textbf{76.8} \\

\midrule
\multicolumn{8}{l}{\textit{\textbf{Profession}}} \\
\quad STEM / Engineering & 30 & 69.0 & 78.8 & 74.0 & 74.5 & 74.8 & \textbf{79.0} \\
\quad Creative / Arts & 18 & 68.5 & 78.5 & 73.0 & 74.2 & 74.5 & \textbf{78.5} \\
\quad Business / Finance & 14 & 67.5 & 77.0 & 70.5 & 74.0 & 73.5 & \textbf{78.0} \\
\quad Student & 15 & 68.5 & 78.5 & 73.5 & 74.5 & 74.5 & \textbf{78.8} \\
\quad Education & 10 & 66.5 & 76.0 & 67.5 & 73.8 & 73.0 & \textbf{77.8} \\
\quad Healthcare & 8 & 66.0 & 75.5 & 66.0 & 73.5 & 72.8 & \textbf{77.5} \\
\quad Manual Trades & 5 & 63.0 & 72.0 & 58.5 & 73.5 & 71.0 & \textbf{76.5} \\

\specialrule{0.8pt}{2pt}{2pt}
\multicolumn{8}{l}{\textit{\textbf{Aggregates}}} \\

\quad Raw aggregate & --- & 67.3 & 77.4 & 70.3 & 74.1 & 73.9 & \textbf{78.3} \\
\quad IPF-reweighted & --- & 66.9 & 73.8 & 64.7 & 73.7 & 72.8 & \textbf{77.4} \\
\quad $\Delta_{\text{IPF}}$ (\%) & --- & $-0.6$ & $-4.7$ & $-8.0$ & $-0.5$ & $-1.5$ & \textbf{$-1.1$} \\

\midrule
\quad $\sigma$ (subgroup) & --- & 2.1 & 2.3 & 5.4 & \textbf{0.3} & 1.2 & \textbf{0.8} \\
\quad Range (max$-$min) & --- & 7.0 & 7.7 & 17.0 & \textbf{1.0} & 4.0 & \textbf{2.5} \\

\bottomrule
\end{tabular}
\end{adjustbox}

\caption{\textbf{Per-subgroup CLIPScore ($\times 100$) on RapidData.}
\MyModel~(5-shot) achieves the highest performance across nearly all subgroups while maintaining low disparity across demographic categories.}

\label{tab:subgroup_cs}
\end{table*}

\paragraph{Age.}
Performance degrades monotonically with age for all data-driven methods, but the rate of degradation differs sharply.
DrUM drops from 73.5~(25--34) to 58.5~(55+), a \textbf{20.4\% decline}---its per-user adapters, trained predominantly on younger users who dominate the interaction data, fail to generalize to older demographics.
TV also degrades~(79.0~$\to$~72.0, $-$8.9\%), as its retrieval pool inherits the same age skew.
\MyModel~(5-shot) degrades gracefully from 79.0 to 76.8~($-$2.8\%), because the persona explicitly encodes age-relevant context (lifestyle, cultural references) that the LLM can leverage regardless of training-data representation.
The unpersonalized LLM baseline itself shows age bias (68.5~$\to$~62.8), consistent with \cite{santurkar2023whose} that language models disproportionately reflect opinions of younger demographics.
ViPer is nearly flat across age groups ($\sigma\!=\!0.4$), which we attribute to its synthetic visual persona training that is decoupled from real demographic distributions; however, its absolute performance remains lower than \MyModel across all groups.

\paragraph{Region.}
The sharpest disparities appear along the regional axis.
DrUM collapses from 75.0~(N.\,America) to 58.0~(Africa), a \textbf{22.7\% drop}---the largest single-subgroup failure we observe.
African, Latin American, and Middle Eastern/CIS users are systematically underserved:
DrUM scores 58.0, 61.5, and 60.0 respectively, all falling in the concerning-to-poor range.
These populations constitute only ${\sim}$15\% of the RapidData sample (\Cref{tab:subgroup_cs}, Sample\,\% column), meaning their poor performance is masked by raw aggregates but surfaced by IPF reweighting.
TV shows similar but milder regional skew~(79.5~$\to$~71.8, $-$9.7\%), while \MyModel~(5-shot) maintains $\geq$76.5 across \emph{all} regions including Africa and ME/CIS.
Even the unpersonalized LLM baseline exhibits Western bias~(69.5~$\to$~62.5), mirroring the English-centric training-data distribution of GPT-4o.

\paragraph{Profession.}
Profession-level patterns echo regional trends.
STEM professionals and students---the groups most overrepresented in crowdsourced AI datasets---receive the strongest personalization from data-driven methods (DrUM: 74.0 and 73.5 respectively).
Manual-trade workers, who constitute only 5\% of the sample, receive the worst treatment: DrUM scores 58.5, TV scores 72.0, and even the LLM baseline drops to 63.0.
We hypothesize that the LLM backbone lacks exposure to the visual vocabulary and aesthetic preferences characteristic of these professions (e.g., industrial photography, trade-specific tools and environments), making it difficult to translate persona cues into effective prompt modifications.
\MyModel~(5-shot) narrows this gap substantially~(79.0~$\to$~76.5), though the 2.5-point residual suggests room for improvement.

\paragraph{PIGReward trends.}
The PIGReward metric, which captures stylistic and cultural alignment beyond CLIP similarity, amplifies the demographic disparities observed in CLIPScore.
\Cref{tab:ipf_summary} reports the IPF degradation for both metrics.
DrUM's PIGReward drops from 63.5~(raw) to 56.8~(IPF), a \textbf{10.6\% decline}---the largest among all methods---confirming that its adapter training absorbs not only the frequency biases but also the stylistic biases of the majority-skewed sample.
TV degrades by 8.0\% on PIG versus 4.7\% on CS, suggesting that retrieval-based methods particularly fail on the nuanced stylistic dimensions (mood, cultural coherence, affect) that PIG captures.
\MyModel~(5-shot) shows minimal PIG degradation~($-$1.4\%), demonstrating that persona conditioning preserves cultural and stylistic alignment equitably.

\begin{table}[t]
\centering
\setlength{\tabcolsep}{4pt}
\renewcommand{\arraystretch}{1.15}
\begin{adjustbox}{max width=\columnwidth}
\begin{tabular}{@{}l l cc c cc c@{}}
\toprule
& & \multicolumn{3}{c}{\textbf{CLIPScore}} & \multicolumn{3}{c}{\textbf{PIGReward}} \\
\cmidrule(lr){3-5} \cmidrule(lr){6-8}
\textbf{Method} & \textbf{Setting}
  & \textbf{Raw} & \textbf{IPF} & \textbf{$\Delta$\%}
  & \textbf{Raw} & \textbf{IPF} & \textbf{$\Delta$\%} \\
\midrule
LLM      & 0-shot & 67.3 & 66.9 & -0.6 & 56.0 & 55.1 & -1.6 \\
TV       & 3-shot & 77.4 & 73.8 & -4.7 & 65.3 & 60.1 & -8.0 \\
DrUM     & FT     & 70.3 & 64.7 & -8.0 & 63.5 & 56.8 & -10.6 \\
ViPer    & FT     & 74.1 & 73.7 & -0.5 & 63.1 & 60.8 & -3.6 \\
\midrule
\MyModel & 0-shot & 73.9 & 72.8 & -1.5 & 61.5 & 60.9 & -1.0 \\
\MyModel & 5-shot & \textbf{78.3} & \textbf{77.4} & \textbf{-1.1} & \textbf{74.0} & \textbf{73.0} & \textbf{-1.4} \\
\bottomrule
\end{tabular}
\end{adjustbox}
\caption{\textbf{IPF degradation summary} on RapidData.
$\Delta$\% denotes the relative change from raw to IPF-reweighted aggregate.
Larger negative values indicate greater degradation on underrepresented subgroups.}
\label{tab:ipf_summary}
\end{table}
\clearpage
\section{User Study}
\label{sec:supp-user-study}

We conduct a two-part human evaluation: (1)~a pairwise preference study measuring whether persona conditioning improves perceived personalization, and (2)~an expert annotation study assessing the accuracy and completeness of mined personas.
The study was approved by the institutional Ethics Review Board (ERB). All participants provided informed consent; no PII is stored beyond the duration of the experiment.

\subsection{Part 1: Pairwise Preference Study}
\label{sec:supp-user-study-pref}

\paragraph{Participants and recruitment.}
We recruited 50 participants from a large research institution, selected to maximize demographic diversity across age (18--55+), gender, profession, and geographic background.

\begin{tcolorbox}[
  enhanced,
  title={\textbf{Participant Intake Questionnaire}},
  fonttitle=\small,
  coltitle=black,
  colbacktitle=gray!15,
  colback=white,
  colframe=gray!60,
  boxrule=0.5pt,
  arc=3pt,
  left=8pt, right=8pt, top=6pt, bottom=6pt,
  fontupper=\small
]
\textbf{\textit{Section A: Demographics}}\\[3pt]
\textbf{Q1.}~What is your age range? \hfill \textcolor{gray}{\scriptsize 18--24\,/\,25--34\,/\,35--44\,/\,45--54\,/\,55+}\\[2pt]
\textbf{Q2.}~What is your gender? \hfill \textcolor{gray}{\scriptsize Male\,/\,Female\,/\,Non-binary\,/\,Prefer not to say}\\[2pt]
\textbf{Q3.}~What is your country or region of residence? \hfill \textcolor{gray}{\scriptsize [Free text]}\\[2pt]
\textbf{Q4.}~What is your current profession or field of study? \hfill \textcolor{gray}{\scriptsize [Free text]}\\[2pt]
\textbf{Q5.}~What is your highest level of education? \hfill \textcolor{gray}{\scriptsize Bachelor’s\,/\,Master’s\,/\,PhD\,/\,Other}\\[6pt]
\textbf{\textit{Section B: Interests \& Hobbies}}\\[3pt]
\textbf{Q6.}~List your top hobbies or leisure activities. \hfill \textcolor{gray}{\scriptsize [Free text, comma-separated]}\\[2pt]
\textbf{Q7.}~What topics do you most frequently read about or discuss? \hfill \textcolor{gray}{\scriptsize [Free text]}\\[6pt]
\textbf{\textit{Section C: Personality \& Lifestyle}}\\[3pt]
\textbf{Q8.}~Would you describe yourself as introverted or extroverted? \hfill \textcolor{gray}{\scriptsize Introverted\,/Slightly Introverted\,/Neither\,/Slightly Extroverted\,/Extroverted}\\[2pt]
\textbf{Q9.}~Choose 3--5 words that best describe your personality. \hfill \textcolor{gray}{\scriptsize [Free text]}\\[2pt]
\textbf{Q10.}~Describe a typical weekend for you in 1--2 sentences. \hfill \textcolor{gray}{\scriptsize [Free text]}
\end{tcolorbox}
\noindent\begin{minipage}{\textwidth}
\captionsetup{hypcap=false}%
\captionof{figure}{User Study Questionnaire for constructing their Persona}\label{lst:user_study_questions}
\end{minipage}

\paragraph{Intake questionnaire.}
Each participant completed a 10-question intake survey designed to capture their demographics, profession, and hobbies---the information needed to construct a natural-language persona.
Crucially, we do \emph{not} ask about visual or aesthetic preferences; the goal is to test whether a persona derived solely from identity and interests can drive effective image personalization.
Responses were fed to GPT-4o to produce a structured natural-language persona for each participant.
The full questionnaire is listed in \Cref{lst:user_study_questions}

\begin{figure}[H]
    \centering
    \includegraphics[width=0.85\linewidth]{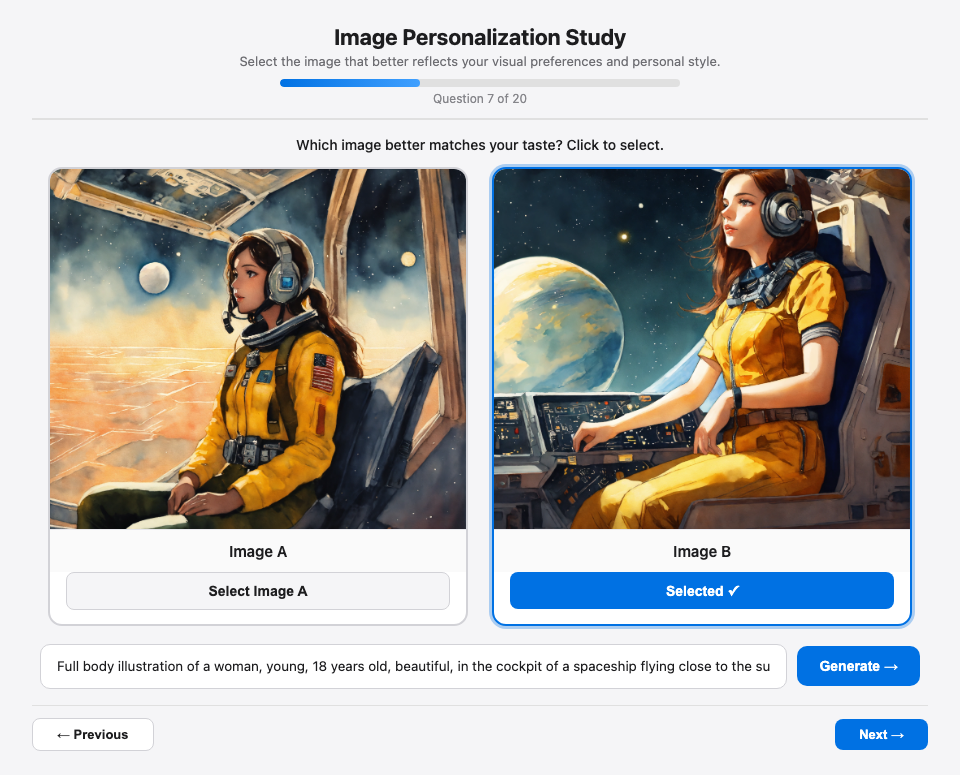}
    \caption{\textbf{Pairwise A/B comparison interface.} Participants enter a prompt and are shown two generated images under different conditions (order randomized). They select the image that better matches their personal visual preferences.}
    \label{fig:user_study_ab_ui}
\end{figure}

\paragraph{Evaluation protocol.}
Using each participant’s persona, we generate images from 20 base prompts under multiple conditions.
Each participant performed pairwise A/B selection over all 20 prompts, choosing the image that better reflected their visual preferences.
The 20 comparisons are split across four conditions (5 prompts each), and the image order (left/right) is randomized per trial:
\begin{enumerate}
    \item \textbf{Persona vs.\ No-Persona} (5 pairs): Zero-shot \MyModel vs.\ unpersonalized LLM rewrite.
    \item \textbf{Own-Persona vs.\ Other-Persona} (5 pairs): \MyModel conditioned on the participant’s own persona vs.\ a randomly assigned other participant’s persona.
    \item \textbf{\MyModel (0-shot and 3-shot, alternating) vs.\ TV (3-shot)} (5 pairs): Our zero-shot persona conditioning vs.\ TV’s 3-shot retrieval baseline.
    \item \textbf{\MyModel (3-shot) vs.\ DrUM / ViPer} (5 pairs each): Few-shot persona conditioning vs.\ fine-tuned baselines (DrUM and ViPer).
\end{enumerate}
The pairwise comparison interface is shown in \cref{fig:user_study_ab_ui}. For each trial, participants enter a text prompt and are shown two images generated under different conditions, labeled only as ``Image A’’ and ``Image B.’’ They select the image that better matches their taste. No method labels or other cues are revealed. This yields $50 \times 20 = 1{,}000$ total annotations.

\paragraph{Results.}
\Cref{tab:user_study_results} summarizes win rates across all four conditions.
In the \textbf{persona vs.\ no-persona} condition, \MyModel achieved a \textbf{79\% win rate}, confirming that persona-aligned outputs are strongly preferred over generic generations.
Against established baselines, \MyModel (0-shot and 3-shot) achieved a 56\% and 78\% win rate over TV (3-shot), and \MyModel (3-shot) achieved 58\% and 65\% win rates over DrUM and ViPer respectively---despite both baselines requiring per-user fine-tuning or extensive interaction histories.

\begin{table}[H]
\centering
\small
\renewcommand{\arraystretch}{1.15}
\setlength{\tabcolsep}{5pt}
\begin{tabular}{@{}lcc@{}}
\toprule
\textbf{Condition (Method A vs.\ B)} & \textbf{Win\% (A)} & \textbf{$n$} \\
\midrule
\MyModel (0-shot) vs.\ No-Persona & \textbf{79\%} & 250 \\
\MyModel (0-shot) vs.\ TV (3-shot) & \textbf{56\%} & 125 \\
\MyModel (3-shot) vs.\ TV (3-shot) & \textbf{78\%} & 125 \\
\MyModel (3-shot) vs.\ DrUM (FT) & \textbf{58\%} & 250 \\
\MyModel (3-shot) vs.\ ViPer (FT) & \textbf{65\%} & 250 \\
\bottomrule
\end{tabular}
\caption{\textbf{User study pairwise preference results.} Win rate indicates the fraction of trials where Method A was preferred. All conditions show significant preference ($p < 0.01$, binomial test) for persona-conditioned outputs. $n$ = number of pairwise comparisons per condition.}
\label{tab:user_study_results}
\end{table}

\subsection{Part 2: Persona Quality Assessment}
\label{sec:supp-user-study-quality}

To validate the fidelity of our persona construction pipeline, we conducted an expert annotation study.
We sampled 100 user profiles from our mined persona bank and recruited three independent expert annotators per profile, yielding approximately 1{,}000 annotations.
Each annotator rated the persona on a 1--5 Likert scale across three criteria: (i) \textit{demographic accuracy} (correctness of inferred age, location, profession), (ii) \textit{interest completeness} (coverage of the user’s behavioral interests), and (iii) \textit{overall coherence} (whether the persona reads as a plausible, self-consistent individual).

Inter-annotator agreement, measured using Cohen’s Kappa, was $\kappa = 0.64$, indicating substantial agreement across reviewers.
The average rating across all criteria was \textbf{4.1 / 5}, with per-criterion scores of 4.3 (demographic accuracy), 3.9 (interest completeness), and 4.1 (overall coherence).
These results confirm that the graph-mined personas are judged to be both accurate and sufficiently complete by human experts, supporting the validity of our persona construction pipeline for downstream personalization.

\section{Limitations}
\label{sec:supp-limitations}
In this section we discuss the limitations of Zero and Few Shot image personalization using Natural Language Personas as proxy for user preferences. We encounter three primary challenges and we discuss how we can mitigate them in future work.
\paragraph{Over Personalization} Highly specific personas (e.g., \textit{111920s Art Deco 
illustrator specializing in geometric patterns and gold-leaf accents"}) occasionally 
override prompt semantics for eg: a request for ``a dog in a park" yields an Art Deco geometric 
dog rather than a naturalistic scene. We empirically observe that setting a lower temperature ($<0.6$) and chain of thought reasoning is able to reduce these to a great extent, we leave these prompt strategies and training methods for future work. \paragraph{Demographic Alignment}: Since \zipbench is built over intersection of Civitai and Reddit, it shows similar demographic skews as other benchmarks like Pick-a-Pic, HPD, PIP. However, we emphasize that by introducing these demographic attributes, we propose the first evaluation framework to measure the equitable demographic alignment of personalization methods. We hope this will steer the community to re-assess their annotation methodology and move towards more aligned benchmarks.
\paragraph{Persona Verbalization Hallucination in long contexts}
We observe that GPT-4o begins 
hallucinating persona details beyond 8192 tokens (fabricating subreddit memberships, 
inventing stylistic preferences), motivating our 4096-token budget cap. We leave more detailed personas and stronger alignment for future work.

\phantomsection
\section*{Ethics, Data Governance, and Responsible Use}
\label{sec:supp-ethics}
\addcontentsline{toc}{section}{Ethics, Data Governance, and Responsible Use}

\paragraph{Public Data and Consent.}
All datasets used in this work are derived exclusively from publicly available Reddit content accessed through the official Reddit API and governed by Reddit’s Terms of Service. No private messages, deleted content, or non-public information are collected. Although Reddit data is public, users may not anticipate research use; therefore, all data handling follows conservative privacy practices. No identifiable metadata (including usernames, timestamps, subreddit combinations, or cross-platform identifiers) is released, and all internal identifiers are randomized and non-reversible.

\paragraph{Anonymization and De-identification.}
Before any modeling or analysis, all user identifiers are hashed, and potentially identifying fields (e.g., usernames, URLs, exact geotags) are removed.

\paragraph{Sensitive Attribute Handling.}
Personas may contain demographic or cultural cues (e.g., ``Brazilian street photographer''), but our pipeline explicitly prohibits extraction, prediction, or use of sensitive attributes such as race, political affiliation, sexual orientation, or health status. Automatic filters and manual review prevent the inclusion of sensitive categories or uniquely identifying personal content. We caution against using persona-driven personalization in any high-stakes context.

\paragraph{Bias, Stereotypes, and Harm Prevention.}
Persona-driven models risk reinforcing stereotypes when demographic or cultural cues appear. To mitigate this, we design prompts that emphasize aesthetic and interest-driven traits rather than sensitive attributes. We audit a sample of model outputs to ensure that stylistic modulation does not drift into culturally insensitive or stereotypical depictions. Further research is required to study fairness, calibration, and user-controlled editing of personas.

\paragraph{Model Behavior and Hallucinations.}
Generative models may hallucinate persona traits, especially with long contexts. To address this, we cap persona token budgets, apply consistency checks, and filter hallucinated or unverifiable claims. Persona descriptions should not be interpreted as factual user profiles, but rather as approximate, interest-oriented abstractions.

\paragraph{Data Release and Reproducibility.}
ZIP-Bench will be released, phased, and in anonymized form with filtered textual content, redacted identifiers, and derisked persona--image pairs. As required by Reddit API policy, raw Reddit posts and comments are not redistributed. We have reached out to individual users for their permission and the public release has 301 users.

\paragraph{Intended Use and Limitations.}
ZIPP is designed for research on personalization, multimodal reasoning, and interpretability, not for commercial profiling, behavioral targeting, or automated inference of private traits. Personas should not be used for decisions about real individuals. Any downstream application should incorporate informed user consent, transparency regarding persona construction, and opt-out mechanisms for personalization.

\end{document}